\documentclass{article}

\usepackage[preprint]{neurips_2025}

\usepackage[utf8]{inputenc} 
\usepackage[T1]{fontenc}    
\usepackage{hyperref}       
\usepackage{url}            
\usepackage{booktabs}       
\usepackage{amsfonts}       
\usepackage{nicefrac}       
\usepackage{microtype}      
\usepackage{xcolor}         

\usepackage{amsmath}
\usepackage{amssymb}
\usepackage{mathtools}
\usepackage{amsthm}
\usepackage{bbm}
\usepackage{multirow}

\theoremstyle{plain}
\newtheorem{theorem}{Theorem}[section]

\newtheorem{lemma}[theorem]{Lemma}

\theoremstyle{definition}
\newtheorem{definition}[theorem]{Definition}
\newtheorem{assumption}[theorem]{Assumption}
\newtheorem{remark}[theorem]{Remark}

\usepackage[textsize=tiny]{todonotes}
\usepackage{enumitem}
\setlist[enumerate]{itemsep=0mm}
\setlist[itemize]{itemsep=0mm}

\usepackage{algorithm}
\usepackage{algorithmic}

\title{Avoiding exp(R) scaling in RLHF through Preference-based Exploration}

\author{%
  Mingyu Chen \\
  Department of Electrical \& Computer Engineering\\
  Boston University\\
  Boston, MA 02215 \\
  \texttt{mingyuc@bu.edu} \\
  \And
  Yiding Chen \\
  Department of Computer Science\\
  Cornell University\\
  Ithaca, NY 14850 \\
  \texttt{yc2773@cornell.edu} \\
  \And
  Wen Sun \\
  Department of Computer Science\\
  Cornell University\\
  Ithaca, NY 14850 \\
  \texttt{ws455@cornell.edu} \\
  \And
  Xuezhou Zhang \\
  Faculty of Computing \& Data Sciences\\
  Boston University\\
  Boston, MA 02215 \\
  \texttt{xuezhouz@bu.edu} \\ \\
}

\begin{document}

\maketitle

\begin{abstract}
Reinforcement Learning from Human Feedback (RLHF) has emerged as a pivotal technique for large language model (LLM) alignment.
This paper studies the setting of online RLHF and focuses on improving its sample efficiency.
All existing algorithms for online RLHF, whether doing passive exploration or active exploration, suffer from a sample complexity that scales exponentially with the range of the reward function.
This statistical inefficiency hinders their effectiveness in scenarios with heavily skewed preferences, e.g. questions with objectively correct answers.
To address this, we introduce \textit{Self-Exploring Preference-Incentive Online Preference Optimization} (\texttt{SE-POPO}), an online RLHF algorithm that for the first time achieves a sample complexity that scales \textit{polynomially} with the reward range, answering an open problem raised by \citet{xie2024exploratory}.
Theoretically, we demonstrate that the sample complexity of \texttt{SE-POPO} dominates that of existing exploration algorithms. Empirically, our systematic evaluation confirms that \texttt{SE-POPO} is more sample-efficient than both exploratory and non-exploratory baselines, in two primary application scenarios of RLHF as well as on public benchmarks, marking a significant step forward in RLHF algorithm design.
\end{abstract}

\section{Introduction}\label{section::introduction}
Reinforcement Learning from Human Feedback (RLHF) has emerged as a pivotal technique in the post-training of Large Language Models (LLMs) \citep{christiano2017deep, ziegler2019fine, ouyang2022training}. 
Earlier works on RLHF focus primarily on the offline setting \citep{ouyang2022training, rafailov2024direct}, where the preference data are pre-collected and fixed prior to the fine-tuning phase.
However, in this setting, the quality of alignment is fundamentally limited by the quality of response in the pre-collected preference dataset.
To overcome this limitation, recent works attempt to perform RLHF in an online setting.
By continually generating and subsequently labeling new samples during training, online RLHF allow the agents to receive feedbacks on out-of-distribution (OOD) responses and achieves improved empirical performance \citep{dong2024rlhf}.

Similar to online reinforcement learning, the most critical challenge in online RLHF is how to balance the \textit{exploration-exploitation trade-off}.  
In naive online RLHF algorithms \citep{guo2024direct}, the exploration is carried out passively, relying solely on the inherent randomness of the LLM policy.  
Such a passive approach fails to sufficiently explore the prompt-response space even with many samples.  
More recently, a number of active exploration algorithms have been proposed \citep{dwaracherla2024efficient, xiong2024iterative, xie2024exploratory, cen2024value, zhang2024self}. 
By leveraging optimism-based approaches to encourage the policy to target OOD regions, active exploration has demonstrated superior performance over passive exploration both in theory and in practice. A more comprehensive discussion on related works is deferred to Appendix \ref{sec:rw}.

However, all existing online RLHF algorithms share one common flaw: They remain effective only when the reward is small. 
In particular, under the Bradley–Terry (BT) model and assuming the reward satisfies $r\in[0,R_{\max}]$, all existing algorithm have a sample complexity in the form of $O(\exp(R_{\max})/\epsilon^2)$, scaling \textit{exponentially} with $R_{\max}$.
Intuitively, this issue arises because human feedback in RLHF is given in the form of preferences instead of numerical rewards. 
Under the BT model, even if there is a significant gap in rewards between two responses, they may behave very similar in their chance of being preferred when pairing with another response that is significantly worse than both. 
As a result, exponentially many samples are necessary to distinguish the quality of responses based on preference signals. 
This leads to the open question raised by \citet{xie2024exploratory}: 
\begin{center}
\textbf{Does there exist an online RLHF algorithm that avoids the exponentially dependency on the reward scale?}
\end{center}

In this work, we answer this question in the positive with a new online RLHF algorithm, \textit{Self-Exploring Preference-Incentive Online Preference Optimization} (\texttt{SE-POPO}), that for the first time achieves a sample complexity that scales \textit{polynomially} with the reward scale.
Our algorithm is provably sample-efficient, scalable and easy to implement.
We summarize our contributions below.

\begin{itemize}[leftmargin=*,topsep=0pt,itemsep=-1ex,partopsep=1ex,parsep=1ex]
    \item We introduce a preference-based exploration technique, distinct from the reward-based exploration done in all prior works.
    Based on this new technique, we design a subroutine algorithm \textit{Preference-Incentive Online Preference Optimization} (\texttt{POPO}), which achieves a preference-based regret that scales polynomially with $R_{\max}$ against a fixed comparator policy.
\item Building upon \texttt{POPO}, we propose a self-sampler update technique that effectively prevents the sample complexity from exploding as reward scale increases. 
Leveraging this idea, we develop our main algorithm \texttt{SE-POPO}, achieving a sample complexity scaling polynomially with $R_{\max}$.
\item
We perform a comprehensive empirical evaluation of our algorithm across multiple training and testing settings as well as on major public benchmarks. In addition, we perform ablation studies to further understand the effect of the sampler update mechanism in our algorithm.
The results show that our algorithm outperforms both exploratory and non-exploratory baselines across all benchmarks with a large margin.
\end{itemize}

\section{RLHF Preliminaries}
In RLHF, we denote a policy by $\pi$, which generates an answer $y\in \mathcal{Y}$ given a prompt $x\in \mathcal{X}$ according to the conditional probability distribution $\pi(\cdot|x)$. 
Given two responses $y$ and $y'$ with respect to prompt $x$, we assume a preference oracle, i.e. a human evaluator, will evaluate the quality of two responses and indicate the preferred one.
Following prior works, we consider Bradley–Terry model as the preference oracle.
The mathematical definition is below.
\begin{assumption}(Bradley–Terry (BT) Model)
\label{ass: BT}
There exists an underlying reward function $r^\star: \mathcal{X}\times \mathcal{Y} \to \mathbb{R} $ such that for every $x,y,y'\in \mathcal{X}\times\mathcal{Y}\times \mathcal{Y}$, 
\begin{align*}
    \mathbb{P}^\star(y\succ y'|x) = \frac{\exp(r^\star(x,y))}{\exp(r^\star(x,y))+ \exp(r^\star(x,y'))} = \sigma(r^\star(x,y)-r^\star(x,y')),
\end{align*}
where $ \mathbb{P}^\star(y\succ y'|x)$ represents the probability that $y$ is preferred to $y'$ given $x$ and $\sigma$ represents the sigmoid function. Without loss of generality, we assume that for all $x,y\in \mathcal{X}\times\mathcal{Y}$, we have $r^\star(x,y)\in [0,R_{\text{max}}]$ and $R_{\text{max}}\ge 1$.
\end{assumption}

\paragraph{The Two-stage RLHF pipeline:}
In the classic two-stage RLHF framework \citep{christiano2017deep, ouyang2022training}, the algorithm assumes access to a dataset $\mathcal{D} = \{x_n, y_n^1, y_n^2, o_t\}_{n=1}^N$, where 
\begin{align*}
    x_n\sim \rho,\ y_n^1\sim \pi_{\text{ref}},\ y_n^2\sim \pi_{\text{ref}},\ o_n\sim \text{Ber}\left( \mathbb{P}^\star(y\succ y'|x)  \right).
\end{align*}
Here, $\rho$ denotes the underlying prompt distribution.
$\pi_{\text{ref}}$ is a reference language model, which is typically obtained via supervised fine-tuning.
$o_n$ is obtained by the preference oracle.
For simplicity, we redefine the dataset as $\mathcal{D} = \{x_n, y_n^w, y_n^l\}_{n=1}^N$, where $y_n^w$ and $y_n^l$ are assigned based on the value of $o_n$.
Given the dataset, we first estimate the reward function via maximum likelihood estimation, i.e.,
\begin{align}
    \hat r= \arg\min_{r\in \mathcal{R}} -\sum_{n=1}^N \log\sigma\left( 
 r(x_n, y_n^w) - r(x_n, y_n^l) \right)
 =: \arg\min_{r\in \mathcal{R}}  \ell(r, \mathcal{D}).\label{rlhf_rm}
\end{align}
With the learned reward function, the objective of RLHF is to fine-tune the policy $\pi$ to maximize the reward.
Following prior theoretical works on RLHF, we consider a KL-regularized reward objective, that is,
\begin{align}
    \hat \pi = \arg\max_{\pi\in \Pi} \mathbb{E}_{x\sim \rho, y\sim \pi(\cdot|x)}\left[\hat r(x,y) - \beta \log\frac{\pi(y|x)}{\pi_{\text{ref}}(y|x)}   \right] 
    =:  \arg\max_{\pi\in \Pi}  J(\hat r, \pi).\label{rlhf_po}
\end{align}
\paragraph{The DPO pipeline:}
An alternative approach of RLHF is introduced by \citep{rafailov2024direct}, namely Direct Preference Optimization (DPO).
The key motivation of DPO is from the closed-form solution of (\ref{rlhf_po}), that is, given a reward function $\hat r$, the solution $\hat \pi$ satisfies
\begin{align}
    \hat \pi (y|x) = \frac{\pi_{\text{ref}(y|x)}\exp(\hat  r(x,y)/\beta)}{Z(r,x)},\ \forall x,y\in \mathcal{X}\times \mathcal{Y}\label{DPO_1}
\end{align}
where $Z(r,x) = \sum_{y} \pi_{\text{ref}(y|x)}\exp(\hat  r(y|x)/\beta)$ is a partition function independent of $y$.
The closed form solution allows us to represent the reward by $\hat \pi$
\begin{align}
    \hat r(x,y) - \hat r(x,y')
    = \beta \log \frac{\hat\pi(y|x)}{\pi_{\text{ref}}(y|x)}-\beta \log \frac{\hat\pi(y'|x)}{\pi_{\text{ref}}(y'|x)}\label{DPO_2}
\end{align}
for every $\forall (x,y,y')\in \mathcal{X}\times \mathcal{Y}\times \mathcal{Y}$.
By substituting (\ref{DPO_2}) into (\ref{rlhf_rm}), DPO bypasses the need for explicitly learning the reward function.
Instead, it optimizes the policy directly with objective
\begin{align}
\hat \pi=\arg\min_{\pi\in \Pi}- \sum_{n=1}^N \log\sigma\left( 
 \beta \log \frac{\pi(y_n^w|x_n)}{\pi_{\text{ref}}(y_n^w|x_n)}-\beta \log \frac{\pi(y_n^l|x_n)}{\pi_{\text{ref}}(y_n^l|x_n)}\right) =: \arg\min_{\pi\in \Pi} \ell(\pi, \mathcal{D}).
 \label{DPO_loss}
\end{align}

\paragraph{Performance metric}
The performance of a learned policy $\hat \pi$ is measured by the suboptimal gap
\begin{align*}
    \text{SubOpt}(\hat\pi) = \mathbb{E}_{x\sim \rho, y\sim \pi^\star(\cdot|x), y'\sim \hat \pi(\cdot|x)}[r^\star(x,y)-r^\star(x,y')],
\end{align*}
where $\pi^\star = \arg\max_{\pi\in \Pi} \mathbb{E}_{x\sim \rho, y\sim \pi^\star(\cdot|x)}[r^\star(x,y)]$ denotes the optimal policy.
Our goal is to propose a sample-efficient and also implementation-friendly algorithm to learn a policy $\hat \pi\in \Pi$ such that $\text{SubOpt}(\hat \pi)\le \epsilon$ for some small $\epsilon>0$.

\paragraph{Online Feedback and Exploration}
In early RLHF studies, the preference dataset $\mathcal{D}$ is typically assumed to be given. 
Although such offline RLHF has been highly successful in aligning language models, it is inherently limited by the quality of the preference data and $\pi_{\text{ref}}$. 
To overcome these limitations, RLHF with online feedback is proposed \citep{guo2024direct}.
In the online framework, the dataset is constructed with human feedbacks on the responses generated from the language model on the fly.
Formally, online RLHF proceeds in $T$ rounds with each round as follows:
\begin{enumerate}[leftmargin=*,topsep=0pt,itemsep=-1ex,partopsep=1ex,parsep=1ex]
    \item The agent computes $\pi_t$ using current data $\mathcal{D}_t$ and samples $x_t\sim \rho, y_t^1\sim \pi_t(\cdot|x), y_t^2\sim \pi_t(\cdot|x)$.
    \item Human labels responses $(x_t, y_t^1, y_t^2)\to (x_t, y_t^w, y_t^l)$. Update $\mathcal{D}_{t+1} = \mathcal{D}_t \cup \{(x_t, y_t^w, y_t^l)\}$.
\end{enumerate}
Although numerous empirical studies have demonstrated the benefits of online RLHF, the theoretical foundation has been missing.
The main reason is that existing methods rely on \textit{passive exploration} to collect data, i.e. the responses are sampled directly from the policy $\pi_t$ relying purely on the randomness of $\pi_t$ for exploration.
Motivated by this, recent works \citep{cen2024value, xie2024exploratory, zhang2024self} start to incorporate the optimism principle into RLHF, which encourages explicitly exploration in the policy $\pi_t$.
Although their implementations differ, the essence of their algorithms is to replace the MLE objectives \eqref{rlhf_rm} and \eqref{rlhf_po} in vanilla RLHF with
\begin{align}
 r_{t+1} =\ \arg\max_{r\in \mathcal{R}} \left\{-\ell
(r, \mathcal{D}_t)+ \alpha J(r, \pi(r))\right \}, 
\text{ s.t. }\pi(r) = \arg\max_{\pi\in \Pi}J(r, \pi) \label{xpo-obj}
\end{align}
where $\alpha \max_{\pi\in \Pi}J(r, \pi)$ is a \textbf{reward-based exploration bonus} that encourages exploration.
Such a bonus leads to an overestimation of rewards with high uncertainty, thereby incentivizing policy to explore uncertain responses.
As shown by \citet{cen2024value, xie2024exploratory}, this design offers a practical and provably sample-efficient online exploration algorithm for RLHF with general function approximation.

\section{Preference-based Exploration}
Although existing algorithms based on \eqref{xpo-obj} obtain theoretical sample efficiency guarantees, there is a significant gap between their bounds and what could be achieved under the standard MDP framework.
In particular, the best known sample complexity bound takes the form of $O(\exp(R_{\max})/\epsilon^2)$, which scales \textit{exponentially} with the reward scale $ R_{\text{max}}$.
This makes existing guarantees quite subtle, as the bound quickly becomes vacuous as soon as $R_{\text{max}}$ is moderately large.
In practical LLM applications, it is common that one response can strictly dominate another, i.e., $ \mathbb{P}^\star(y \succ y' | x) \to 1 $.
Under the BT model (Asm.~\ref{ass: BT}), this implies a very large $R_{\text{max}}$. Authors of prior works have admitted that this is a significant drawback of these results and in fact conjectured that the exponential dependency might be unavoidable \citep{xie2024exploratory}. In this paper, we resolve this conjecture in the negative by presenting the first algorithm that avoids such exponential dependency on the reward scale.
In what follows, we start by discussing the cause of exponential dependency on $R_{\text{max}}$ and why it's a real limitation of the algorithms rather than merely an artifact of the analysis. After that, we will present our technique that solves it.

\subsection{The cause of $\mathbf{exp(R_{max})}$ scaling}\label{section:1}
Using online-to-batch technique, the sample complexity of an online algorithm can be derived from its regret, which is defined by $\sum_{t=1}^T \text{SubOpt}(\pi_t)$.
In the standard analysis of optimism online RLHF, the regret can be bounded by the sum of reward uncertainty, i.e., $\sum_{t=1}^T \mathbb{E}_{x\sim \rho, y\sim \pi_t(\cdot|x)}[|r_t(x,y)-r^\star(x,y)|]$, where $r_t$ is the induced reward function from $\pi_t$ as in DPO.
To bound the reward uncertainty, prior works reduce it to the preference uncertainty, i.e., $\sum_{t=1}^T \mathbb{E}_{x \sim \rho, y \sim \pi_t(\cdot | x), y' \sim \pi_{t}(\cdot | x)} [|\mathbb{P}_t(y \succ y' | x) - \mathbb{P}^\star(y \succ y' | x)|]$, as the preference uncertainty can be effectively bounded using concentration inequalities.
Unfortunately, this reduction is not a free lunch: due to the presence of sigmoid function in Bradley–Terry Model, for some $x,y,y'$, there is
\begin{align}
    |r_t(x,y)-r^\star(x,y)|\approx \frac{|\mathbb{P}_t(y \succ y' | x) - \mathbb{P}^\star(y \succ y' | x)|}{\nabla \sigma(r^\star(x,y)-r^\star(x,y'))}
    \label{p2r-exp}
\end{align}
Therefore, the reward uncertainty could be of order $ 1 / \nabla \sigma(R_{\text{max}})\approx \mathcal{O}(\exp(R_{\text{max}})) $ times the preference uncertainty, in the worst case where the reward gap between the two responses $y$ and $y'$ is large. Similar issues have also been discovered in logistic bandits \citep{faury2020improved}.
This explains where $\exp(R_{\text{max}})$ comes from in the theoretical analysis of existing works and highlights the key question in algorithm design: \textbf{How should we sample the responses $y$ and $y'$ in online RLHF?}
A number of prior works \citep{xiong2024iterative, dong2024rlhf, shi2024crucial} use $\pi_t$ to sample $y_t^1$ and use $\pi_{\text{ref}}$, or a policy distinct from $\pi_t$, to sample $y_t^2$. This destines to perform poorly due to \eqref{p2r-exp}. In general, sampling $y_t^2$ using an underperformed policy, such as $\pi_{\text{ref}}$ implies that the reward gap $r^\star(x_t, y_t^1)-r^\star(x_t, y_t^2)$ would be relatively large, causing $ y_t^1 $ to be consistently favored, even if $ y_t^1$ itself is suboptimal.
As a result, such algorithms will struggle to learn the optimal response, as such a comparison provides very little information on how to improve based on the current best policy $\pi_t$.

\subsection{Algorithm Design}
Given the above intuition, we propose \textit{Preference-Incentive Online Preference Optimization with Self-updated Sampler} (\texttt{SE-POPO}), which for the first time enjoys a sample complexity bound that scales \textbf{polynomially with $R_{\text{max}}$}.
Conceptually, \texttt{SE-POPO} differs from prior algorithms in two main aspects: 1) it uses a preference-based exploration bonus instead of a reward-based bonus to explore more efficiently, and 2) it updates the second sampler at intervals instead of fixing it as $\pi_{\text{ref}}$, bypassing the design flaw discussed above.
The pseudocode of the algorithms is presented in Algorithm~\ref{alg:SE-POPO} and \ref{alg:POPO}.

\texttt{SE-POPO} operates over $K$ intervals.
In each interval, \texttt{SE-POPO} selects a fixed sampler $\pi_{\text{sam}}$ to generate the second response and runs the subroutine \texttt{POPO} for $T$ iterations.
The output of \texttt{POPO} is used as the sampler for the next interval, and the output from the last interval serves as the output of \texttt{SE-POPO}.
Let us now present the subroutine \texttt{POPO}.
As illustrated in Algorithm~\ref{alg:POPO}, \texttt{POPO} shares a similar structure with existing optimism RLHF algorithms \citep{xie2024exploratory, zhang2024self, cen2024value}.
However, unlike prior designs that are tailored towards bounding the \textbf{reward-based regret}, i.e. $\sum_{t=1}^T \text{SubOpt}(\pi_t)$, \texttt{POPO} takes an indirect approach and instead optimize the \textbf{preference-based regret} over a fixed sampler $\pi_{\text{sam}}$: 
\begin{align}\label{eq:ref_reg}       \text{Reg}_{\text{pref}}(\pi_{\text{sam}},T)\coloneqq \sum_{t=1}^T\underset{ 
 \substack{x\sim \rho,  (y^\star, y, y')\sim \pi^\star\otimes\pi_t \otimes \pi_{\text{sam}}(\cdot|x)}}{\mathbb{E}}\bigg[\mathbb{P}^\star(y^\star\succ y'|x)-\mathbb{P}^\star(y\succ y'|x)  \bigg].
\end{align}
To achieve this, \texttt{POPO} optimizes the following objective function instead of \eqref{xpo-obj}:
\begin{align}
 r_{t+1} =\ &\arg\max_{r\in \mathcal{R}} \left\{-\ell
(r, \mathcal{D}_t)+ \alpha G(r, \pi(r))I(r, \mathcal{D}_t)\right \}, \text{ s.t. }\pi(r) = \arg\max_{\pi\in \Pi}J(r, \pi)
\label{popo-obj}
\end{align}
Here, $G$ is the expected preference rate of $\pi$ over $\pi_{\text{sam}}$
\begin{align*}
    G(r, \pi) = \mathbb{E}_{x\sim \rho, y\sim \pi(\cdot|x), y'\sim \pi_{\text{sam}}(\cdot|x)}\left[\mathbb{P}_{r}\left(y\succ y'|x\right)  \right],
\end{align*}
where $\mathbb{P}_{r}$ denotes the preference oracle parameterized by reward $r$.
$I(\mathcal{D}_t)$ is an indicator function 
\begin{align*}
   I(r, \mathcal{D}_t) = \mathbbm{1}\left\{\ell(r,\mathcal{D}_t)-\ell(\bar r,\mathcal{D}_t)\le \gamma\right\},
\end{align*}
where $\bar r = \arg\min_{r\in \mathcal{R}} \ell(r, \mathcal{D}_t)$ represents the MLE-based reward estimator.
In brief, \texttt{POPO} applies a truncated preference-based exploration bonus on the reward learning objective.
This design ensures that optimistic exploration is conducted directly with respect to preferences rather than rewards, while also constraining the exploration to regions near the current MLE estimator, thereby mitigating the risk of over-exploration.

Assuming that \texttt{POPO} achieves low preference-based regret, let's now look at how \texttt{POPO} with self-updated samplers eliminates the exponential dependence on $ R_{\text{max}} $ in the reward-based regret.
The key observation is a novel \textit{Preference-to-Reward} reduction lemma as follows.
\begin{lemma}(Preference-to-Reward reduction)
Given any prompt $x\in \mathcal{X}$, let $y^\star$ denotes the optimal response $y^\star = \arg\max_{y\in \mathcal{Y}} r^\star(x,y)$.
For every $(y,y')\in \mathcal{Y}\times\mathcal{Y}$, there is $\mathbbm{1}\{r^\star(x,y)-r^\star(x,y')\le 1\}[ r^\star(x,y^\star)-r^\star(x,y)] 
   \le 20 R_{\text{max}}\left[ \mathbb{P}^\star(y^\star \succ y'|x)-  \mathbb{P}^\star(y\succ y'|x) \right]$.
\label{Pre2Rwd-reduction}
\end{lemma}
The proof of Lemma~\ref{Pre2Rwd-reduction} is deferred to the appendix.
Intuitively, Lemma~\ref{Pre2Rwd-reduction} tells us that the exponential blow-up in preference-to-reward reduction only occurs when $ r^\star(x, y) - r^\star(x, y') $ is large. 
Assuming $ y' \sim \pi_{\text{sam}}(\cdot | x) $. 
If $ \pi_{\text{sam}} $ is ``good enough'' such that $ r^\star(x, y) -r^\star(x, y')\le 1 $ holds for all $x$, we can easily bound the reward-based regret by $ \text{Reg}_r(T) \le \mathcal{O}(R_{\max}) \text{Reg}_{\text{pref}}(\pi_{\text{sam}}, T) $, and thus get rid of the exponential dependence on $R_{\text{max}}$.
So how do we find a good enough sampler \( \pi_{\text{sam}} \)? 
An intuitive idea is to first run \texttt{POPO} to find a suboptimal policy, then use this policy as $\pi_{\text{sam}}$ and rerun \texttt{POPO}.
However, notice that finding a good enough policy by running \texttt{POPO} from scratch would still requires \(\mathcal{O}(\exp(R_{\text{max}}))\) iterations, as we would have been using \( \pi_{\text{ref}} \) as the sampler, and \( \pi_{\text{ref}} \) might be $O(R_{\text{max}})$ worse than $\pi^*$.
The trick, as shown in Algorithm~\ref{alg:SE-POPO}, is to repeat the \texttt{POPO} subroutine for many times and gradually improve $\pi_{\text{sam}}$. 
The main observation is that even if the sampler performs poorly, \texttt{POPO}'s output policy can still achieve a reward higher by a constant amount compared to the sampler.
For instance, consider \( x, y^\star, y' \) such that \( r^\star(x, y^\star) - r^\star(x, y') \) is large. 
If we use \( y' \) as the second response, after \( T \) iterations, we can find a \( y \) such that \( P^\star(y \succ y' | x) \ge P^\star(y^\star \succ y' | x) - \tilde{\mathcal{O}}(1/\sqrt{T}) \) by the preference-based regret (\ref{eq:ref_reg}). 
Since \( r^\star(x, y^\star) - r^\star(x, y') \) is large, \( P^\star(y^\star \succ y' | x) \) will be close to $1$, resulting in \( P^\star(y \succ y' | x) \) being significantly greater than $1/2$, which implies that there is a constant improvement between \( r^\star(x, y) \) and \( r^\star(x, y') \).
Therefore, by repeating \texttt{POPO} $K=\mathcal{O}(R_{\text{max}})$ intervals, the sampler will finally become sufficiently effective.

\begin{algorithm}[tb]
   \caption{SE-POPO: Self-Exploring Preference-Incentive Online Preference Optimization}\label{alg:SE-POPO}
\begin{algorithmic}
   \STATE {\bfseries Input:} Reference policy $\pi_{\text{ref}}$, Policy set $\Pi$, Iterations $T$, Intervals $K$
   \STATE Initialize $\pi_{\text{sam}}^1 \gets \pi_{\text{ref}}$.
   \FOR{$k=1,\dots, K-1$}
   \STATE Update the sampler $\pi_{\text{sam}}^{k+1}\gets \texttt{POPO}(\pi_{\text{ref}}, \pi_{\text{sam}}^{k},\Pi,T)$.
   \ENDFOR
   \STATE Return policy $\bar \pi = \texttt{POPO}(\pi_{\text{ref}}, \pi_{\text{sam}}^{K}, \Pi,T)$.
\end{algorithmic}
\end{algorithm}

\begin{algorithm}[tb]
\caption{POPO: Preference-Incentive Online Preference Optimization}\label{alg:POPO}
\begin{algorithmic}
   \STATE {\bfseries Input:} Reference policy $\pi_{\text{ref}}$, Sampler $\pi_{\text{sam}}$, Policy set $\Pi$, Iterations $T$
   \STATE Initialize $\pi_1 = \pi_{\text{ref}}$.
   \FOR{$t=1,\dots,T$}
   \STATE Generate data $x_1\sim \rho, y_t^1\sim \pi_t(\cdot|x), {y}_t^2\sim \pi_{\text{sam}}(\cdot|x)$.
   \STATE Label the two responses: $(x_t, y^1_t, y^2_t)\to (x_t, y_t^w, y_t^l)$.

    \STATE Optimize objective (\ref{popo-obj-dpo-theory}). Get $\pi_{t+1}$.

   \ENDFOR
   \STATE Return policy $\bar \pi = \text{Uniform}(\pi_1,\dots, \pi_t)$.
\end{algorithmic}
\end{algorithm}

\subsection{Implementation-friendly Objective}
\label{sec:implementation-friendly}
Similar to that of vanilla two-stage RLHF, \eqref{popo-obj} is a bilevel optimization involving both reward and policy, and is challenging to solve in practice. 
Fortunately, $\pi(r)$ remains to be the solution to the KL-regularized reward optimization objective, therefore (\ref{DPO_2}) continues to hold.
By substituting (\ref{DPO_2}) into (\ref{popo-obj}), similar to what is done in \texttt{DPO}, we can bypass the reward model and directly optimize the policy.
Therefore, the objective can be rewritten as
\begin{align}
    \pi_{t+1} = \arg\max_{ \pi\in \Pi} \left\{ -\ell(\pi, \mathcal{D}_t) + \alpha G(\pi) I(\pi, \mathcal{D}_t) \right\},
    \label{popo-obj-dpo-theory}
\end{align}
where $\ell(\pi, \mathcal{D}_t)$ is the DPO loss as in \eqref{DPO_loss}. $G(x)$ is the exploration bonus defined by
\begin{align*}
    G(\pi) = \underset{ 
 \substack{x\sim \rho,y\sim \pi(\cdot|x),y'\sim \pi_{\text{sam}  }(\cdot|x)}}{\mathbb{E}}\left[  \sigma\bigg(\beta \log \frac{\pi(y|x)}{\pi_{\text{ref}}(y|x)} - \beta \log \frac{\pi(y'|x)}{\pi_{\text{ref}}(y'|x)}\bigg)  \right],
\end{align*}
and $I(\pi, \mathcal{D}_t)= \mathbbm{1}\left\{\ell(\pi,\mathcal{D}_t)-\ell(\bar \pi,\mathcal{D}_t)\le \gamma\right\}$ with $\bar \pi = \arg\min_{\pi\in \Pi} \ell(\pi, \mathcal{D}_t)$.
In addition, evaluating $I(r, \mathcal{D}_t)$ requires pre-computing the MLE estimator $\bar \pi$ first, which doubles the computation cost.
In our experiments, we find that the truncation $I(r, \mathcal{D}_t)$ is rarely active and can therefore be omitted. These steps result in the implementation-friendly objective below
    \begin{align}
        \pi_{t+1} = \arg\max_{ \pi\in \Pi} \sum_{s=1}^t  &\log \sigma  \bigg(\beta \log \frac{\pi(y^w_s|x_s)}{\pi_{\text{ref}}(y^w_s|x_s)} - \beta \log \frac{\pi(y^l_s|x_s)}{\pi_{\text{ref}}(y^l_s|x_s)}\bigg) \nonumber\\
        &+ \alpha \underset{ 
 \substack{x\sim \rho,y\sim \pi(\cdot|x),y'\sim \pi_{\text{sam}  }(\cdot|x)}}{\mathbb{E}}\left[  \sigma\bigg(\beta \log \frac{\pi(y|x)}{\pi_{\text{ref}}(y|x)} - \beta \log \frac{\pi(y'|x)}{\pi_{\text{ref}}(y'|x)}\bigg)  \right].
        \label{popo-obj-dpo}
    \end{align}
On paper, \eqref{popo-obj-dpo} can already be implemented efficiently into existing online DPO pipeline \cite{guo2024direct} with a one-line change of the code. 
However, one challenge we encounter when implementing \eqref{popo-obj-dpo} is that calculating the gradient of the objective function requires sampling new responses $y\sim \pi(\cdot|x)$. 
While such sampling is techniquely feasible, we empirically found that this on-policy sampling step is extremely slow in language model finetuning due to the lack of efficient LLM online inference libraries.
To bypass this issue, we decide to prune the first term within the bonus all together, resulting in the following objective:
\begin{align}
        \pi_{t+1}=\arg\max_{ \pi\in \Pi} \sum_{s=1}^t & \log \sigma  \bigg(\beta \log \frac{\pi(y^w_s|x_s)}{\pi_{\text{ref}}(y^w_s|x_s)} - \beta \log \frac{\pi(y^l_s|x_s)}{\pi_{\text{ref}}(y^l_s|x_s)}\bigg) \nonumber\\
        &+ \alpha \underset{ 
 \substack{x\sim \rho\\y'\sim \pi_{\text{sam}  }(\cdot|x)}}{\mathbb{E}}\left[  \sigma\bigg( - \beta \log \frac{\pi(y'|x)}{\pi_{\text{ref}}(y'|x)}\bigg)  \right].\label{popo-obj-dpo-r}
\end{align}
Surprisingly, objective (\ref{popo-obj-dpo-r}) still yields in a sample-efficient algorithm in theory.
We defer further discussion on (\ref{popo-obj-dpo-r}) to Appendix~\ref{sec:lightweight} and now move on to presenting our main theoretical results.
\subsection{Theoretical Guarantees}
\label{sec:theory}
Let the regularization parameter $\beta>0$ be fixed. We start by a \textit{reward realizability} assumption, which states that the reward class used in \texttt{SE-POPO} is sufficiently expressive.
\begin{assumption}(Reward realizability)
\label{assumption:1}
    There exists a set of reward functions $\mathcal{R}$ satisfying $r^\star\in \mathcal{R}$.
\end{assumption}
Given Assumption~\ref{assumption:1}, we define $\mathcal{P}$ as the set of preference model induced by $\mathcal{R}$, and define $\Pi$ as the optimal policies induced by $\mathcal{R}$ under KL-regularized reward objective (\ref{rlhf_po}).
Notice that $|\mathcal{P}| = |\mathcal{R}| = |{\Pi}|$ by definition.
For ease of understanding, we will present our main theorems under the linear reward model setting and defer results for general function approximation to Appendix~\ref{appendix:generalization_bl}.
\begin{assumption} (Linear reward oracle)
\label{assumption:2}
    Every reward $r\in \mathcal{R}$ can be parameterized by
    \begin{align*}
        r_\theta(x, y) = \langle \phi(x,y), \theta  \rangle,\ \forall (x,y)\in \mathcal{X}\times \mathcal{Y},
    \end{align*}
    where $\phi(x,y): \mathcal{X}\times \mathcal{Y}\to \mathbb{R}^d$ is a fixed feature mapping and $\theta\in \mathbb{R}^d$ is the parameter. Without loss of generality, we further assume that $|\phi(x,y,y')|\le 1$ for all $x,y,y'$ and $\|\theta\|_2\le R_{\text{max}}$.
\end{assumption}
The following is the preference-based regret bound for \texttt{POPO}.
\begin{theorem}
\label{theo:POPO}
Given Assumption~\ref{assumption:1} and \ref{assumption:2}, setting $  \alpha = \sqrt{\frac{d\log {T}/{d}}{R_{\text{max}}T\log {|\mathcal{R}|}/{\delta} }}$ and $\gamma = 2\log \frac{|\mathcal{R}|}{\delta}$, with probability $1-2\delta$,
\texttt{POPO} output a policy $\bar \pi$ such that
\begin{align*}
    \underset{ 
 \substack{x\sim \rho\\  (y^\star, y, y')\sim \pi^\star\otimes\bar \pi \otimes \pi_{\text{sam}}(\cdot|x)}}{\mathbb{E}}\bigg[\mathbb{P}^\star(y^\star\succ y'|x)-\mathbb{P}^\star(y\succ y'|x)  \bigg]\le \tilde{{\mathcal{O}}}\left( \sqrt{\frac{{dR_{\text{max}}\log\frac{|\mathcal{R}|}{\delta}}}{T}} + \beta  C_{\text{KL}}  \right)
\end{align*}
    where $C_{\text{KL}} = \mathbb{E}_{x\sim \rho}\left[ 
\mathbb{D}_{\text{KL}}(\pi^\star(\cdot|x)||\pi_{\text{ref}}(\cdot|x))  \right]$.
\end{theorem}

Theorem~\ref{theo:POPO} established a clean $\tilde{\mathcal{O}}(\sqrt{dT})$ bound on the preference-based regret. 
This implies, for example, if one were to train against a strong baseline $\pi_{\text{sam}}$, e.g. GPT-4o, $\texttt{POPO}$ would achieve a winrate against GPT-4o similar to that of the optimal policy with a fast rate of convergence. Of course, in practice, we may not have such strong baselines at our disposal. $\texttt{SE-POPO}$ is designed to achieve a similar performance even without such baselines, by iteratively updating its $\pi_{\text{sam}}$. Our main theorem is presented as follows.
\begin{theorem}
\label{theo:SE-POPO}
Assuming $C_{\text{KL}}$ is well-bounded.
Setting $K=\lceil R_{\text{max}}\rceil $, with probability $1-\delta$, 
\texttt{SE-POPO} output a policy $\bar \pi$ such that
\begin{align*}
    \underset{ 
 \substack{x\sim \rho\\  (y^\star, y)\sim \pi^\star\otimes\bar \pi (\cdot|x)}}{\mathbb{E}}\bigg[r(x, y^\star) - r(x, y)  \bigg]\le \tilde{{\mathcal{O}}}\left( \sqrt{\frac{{dR_{\text{max}}^8\log\frac{|\mathcal{R}|}{\delta}}}{N}} + \beta R_{\text{max}}^3 C_{\text{KL}}  \right).
\end{align*}
Specifically, with $\beta = o({1}/{\sqrt{T}})$, \texttt{SE-POPO} outputs $\epsilon$-optimal policy with $\tilde{\mathcal{O}}\left(\frac{{dR_{\text{max}}^8\log\frac{|\mathcal{R}|}{\delta}}}{\epsilon^2}\right)$ samples.
\end{theorem}

\begin{remark}
Theorem~\ref{theo:SE-POPO} offers a significant improvement over all prior sample complexity bounds for RLHF algorithms under the BT-model, being the first sample complexity bound that scales polynomially with $R_{\text{max}}$.
Compared to prior works on online RLHF \citep{das2024provably, rosset2024direct, xie2024exploratory,zhang2024self, cen2024value}, Theorem~\ref{theo:SE-POPO} retains the same dependencies on the coverage parameter $d$ and precision $\epsilon$, while successfully eliminating the exponential dependence on $R_{\text{max}}$ and $1/\beta$.
Furthermore, in Appendix~\ref{appendix:generalization_bl}, we demonstrate that the theoretical results of \texttt{POPO} and \texttt{SE-POPO} can be generalized beyond linear preference oracle using a general complexity measure proposed in \citep{zhong2022gec}, extending our theoretical results to the general function approximation setting.
\end{remark}

\section{Experiments}
\begin{table*}[ht]
\centering
{%
\resizebox{1.0\linewidth}{!}{
\begin{tabular}{@{}lccccccc@{}}
\toprule
\textbf{Model} & \multicolumn{2}{c}{\textbf{IID Data}} & \multicolumn{2}{c}{\textbf{Alpaca Data}} & \multirow{2}{*}{\textbf{AE2 LC}} & \multirow{2}{*}{\textbf{MT-Bench}}& \multirow{2}{*}{\textbf{Avg. Len. (in AE2)}} \\
\cmidrule(lr){2-3} \cmidrule(lr){4-5}
 & WR & AvgR & WR & AvgR &  &  \\
\midrule

Llama-3-8B-SFT & - & - & 29.5 & 71.57 & 10.20 & 7.69 & 1182\\ 
\midrule
DPO-iter1 & 62.4 & -4.50 & 78.1 & -6.02 & - & - & 1645\\
DPO-iter2 & 66.6 & -3.59 & 87.1 & -3.34 & - & - & 2045\\
DPO-iter3 & 72.4 & -2.33 & 91.3 & -0.02 & 36.10 & 8.28 & 2257 \\
\midrule
XPO-iter1 & 62.6 & -4.40 & 78.3 & -5.79 & - & - & 1674\\
XPO-iter2  & 67.3 & -3.28 & 88.0 & -2.60 & - & - & 2200 \\
XPO-iter3 & 73.0  & -2.09 & 91.8 & 0.60 & 38.23 & 8.21 &2346 \\
\midrule
SE-POPO-iter1  & 62.5 & -4.32 & 80.0 & -5.68 & - & - & 1797\\
SE-POPO-iter2  & {68.2} & -3.15 & 89.1 & -2.45 & - & - & 2302\\
SE-POPO-iter3  & {\textbf{73.3}} & {\textbf{-2.03}} & {\textbf{92.4}} & {\textbf{0.61}} & {\textbf{40.12}} & {\textbf{8.39}} & 2358\\
\midrule
Llama-3-8B-Instruct  & 48.4 & -6.77 & 87.0 & -3.42 & 22.92 & 8.16 & 1899\\
    Llama-3-405B-Instruct    & - & - & - & - & 39.30  &- & 1988 \\

\bottomrule
\end{tabular}%
}
}
\caption{Performance comparison across multiple chat benchmarks.}
\label{tab:comparison}
\end{table*}
In this section, we provide a comprehensive empirical evaluation of \texttt{SE-POPO} in LLM alignment tasks. There are two primary use cases for LLM alignments in real practices:
\begin{enumerate}[leftmargin=*,topsep=0pt,itemsep=-1ex,partopsep=1ex,parsep=1ex]
    \item \textbf{Domain-specific alignment}: This is where the goal is to fine-tune LLMs for a specific type of task, e.g. fashion design.
    \item \textbf{Generalist algnment}: This is where the goal is to train a general-purpose question answering AI that could answer a wide variety of questions. This is for instance what GPTs are designed for.
\end{enumerate}

Importantly, in both use cases, the preference feedback during both training and evaluation would have been provided by \textbf{the same oracle}, e.g. human evaluators. In other words, there should not be any distribution shift in the underlying preference model between training and testing. What distinguishes the two use cases is the prompt distribution during training and deployment. For use case 1, the prompts should come from the same domain during both training and deployment, i.e. no distribution shift in the prompt distribution. For use case 2, the prompt distribution between training and testing could be different.

Motivated by the real use cases discussed above, we present three sets of experiments.
For all experiments, 
our implementation build upon the iterative \texttt{DPO} codebase from \citep{dong2024rlhf}, and we use the $3$-iteration online RLHF framework following the setting in \citep{xie2024exploratory}.
Across all three experiments, we use \texttt{Llama-3-8B-SFT} as the base model, \texttt{RLHFlow-ultrafeedback} dataset as the training prompt sets, and \texttt{GRM-Llama3-8B-rewardmodel-ft} as the training preference model. More details about the experiment setup are deferred to Appendix~\ref{appendix-exp}.
The results from the three sets of experiments are shown as three columns in Table~\ref{tab:comparison}:
\begin{itemize}
[leftmargin=*,topsep=0pt,itemsep=-1ex,partopsep=1ex,parsep=1ex]
    \item \textbf{``IID data"} refers to the setting where the models are evaluated on a held-out test prompt set that are drawn from the same distribution as the training prompt set, and the responses are evaluated by the same preference model used during training. This is to simulate Use Case \#1.
    \item \textbf{``Alpaca data"} refers to the setting where the models are evaluated on the AlpacaEval 2.0 dataset, but the responses are still evaluated by the same preference model used during training. This is to simulate Use Case \#2.
    \item \textbf{Public benchmarks}: Finally, we also evaluate our algorithm on public benchmarks including AlpacaEval 2.0 and MT-bench shown in Table~\ref{tab:comparison} as well as the academic benchmarks that are deferred to Table~\ref{tab:academic} in the appendix. These public benchmarks all have one common characteristic: the training and evaluation preference models are different, usually with GPT-4o as the evaluation oracle during testing. As discussed above, such a distribution shift in the preference model between training and testing rarely happen in practice. Thus, we believe that \textbf{the performances on such benchmarks offer little insight on how well an RLHF algorithm works in practice}. Nevertheless, we include them for completeness due to their wide adoption in prior RLHF research.
\end{itemize}

\textbf{Baselines: } 
We compare against two baseline algorithms: iterative \texttt{DPO} \citep{dong2024rlhf}, which is the state-of-the-art passive exploration algorithm and \texttt{XPO} \citep{xie2024exploratory} which is the state-of-the-art active exploration algorithm.
Importantly, here we use the practical implementation of \texttt{XPO}, where both responses are drawn from the previous policy $\pi_t$, rather than from $\pi_t$ and the reference policy $\pi_{\text{ref}}$.
We defer the results of the theoretical \texttt{XPO} to Appendix~\ref{appendix-exp}.
Empirically, the practical implementation of \texttt{XPO} significantly outperforms its theoretical version presented in the paper's main text.

\textbf{Results: } As can been seen in Table \ref{tab:comparison},
\texttt{SE-POPO} outperforms both \texttt{DPO} and \texttt{XPO} across all experiment setups.
Moreover, on the public benchmarks, \texttt{SE-POPO} achieves better performance compared to the industry-level 8B model (Llama-3-8B-Instruct) and comparable performance to model with two orders of magnitude more parameters (Llama-3-405B-Instruct).
Beyond instruction-following benchmarks, we also evaluate \texttt{SE-POPO} and the baselines on a suite of academic benchmarks, to demonstrate that our improvements in chat capabilities do not come at an additional expense of reasoning ability compared to other baselines.
The results are deferred to Appendix~\ref{appendix-exp}.
Across the $9$ academic tasks evaluated, our algorithm performs best in $4$, while DPO leads in $3$ and XPO in $2$.
These evaluation results resoundingly support the effectiveness of our algorithm.

\textbf{Slight length exploitation in \texttt{XPO} and \texttt{SE-POPO}:} 
It is worth noting that the length of the responses generated with models trained by \texttt{XPO} and \texttt{SE-POPO} are slightly longer compared to \texttt{DPO}. This makes sense in theory, considering that the exploration term in both \texttt{XPO} loss and (\ref{popo-obj-dpo-r}) encourages minimizing \( \log \frac{\pi(y'|x)}{\pi_{\text{ref}}(y'|x)} \), which inherently incentives models to generate longer responses.
We speculate that using objective (\ref{popo-obj-dpo}) can mitigate this exploitation, as the on-policy term \(\log \frac{\pi(y|x)}{\pi_{\text{ref}}(y|x)} \) in (\ref{popo-obj-dpo}) will encourage \( \pi \) to generate shorter responses, thereby counteracting the effect incurred by \( \log \frac{\pi(y'|x)}{\pi_{\text{ref}}(y'|x)} \).
Unfortunately, we cannot implement the version of \texttt{SE-POPO} with objective (\ref{popo-obj-dpo}) and have to defer a more comprehensive study of this phenomenon to future work.

\begin{figure*}[t]
\begin{minipage}[c]{0.36\linewidth}
        \centering
        \begin{tabular}{l|cc}
        \toprule
        \textbf{Model}  &\textbf{WR} & \textbf{AvgR}  \\
        \midrule
        $(\pi_t, \pi_t)$-iter2       & 87.0 & -3.35  \\
         $(\pi_t, \pi_t)$-iter3       & 91.2 & -0.02 \\
        \midrule
        $(\pi_t, \pi_{\text{ref}})$-iter2 & 86.8 & -4.09 \\
       
        $(\pi_t, \pi_{\text{ref}})$-iter3 & 89.4 & -2.63\\
        \bottomrule
        \end{tabular}
         \caption{Avg. Reward and Win Rate Comparison.}
        \label{tab:2}
\end{minipage}
\begin{minipage}[c]{0.68\linewidth}
        \centering        \includegraphics[width=0.48\linewidth]{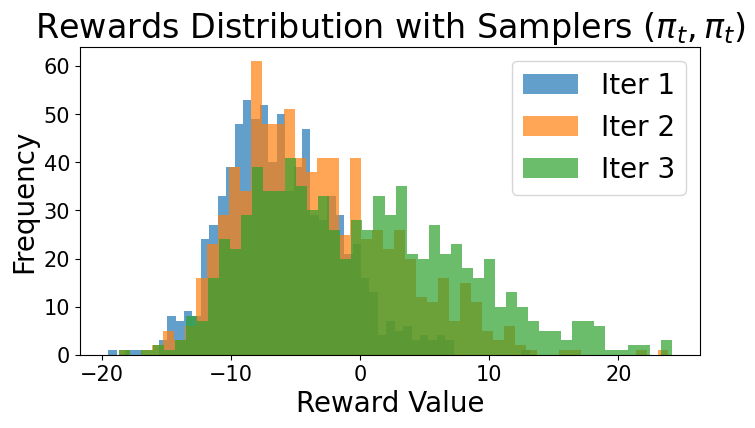}
        \includegraphics[width=0.48\linewidth]{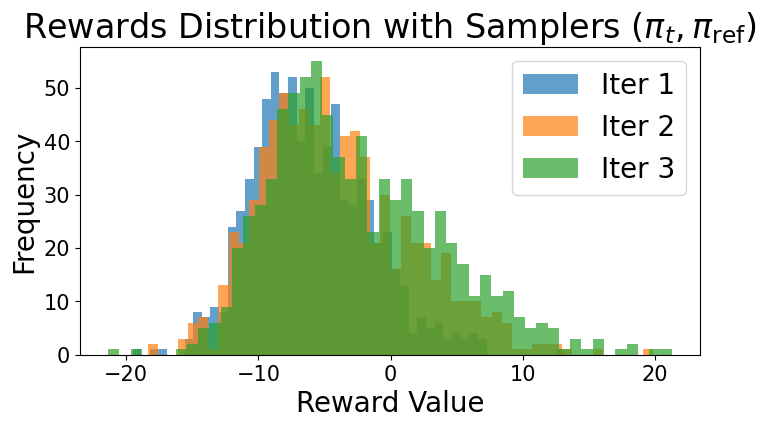}
    \caption{Rewards Distribution with Different Samplers.}
    \label{fig:img1}    
    \end{minipage}
    \vspace{-10pt}
\end{figure*}

\textbf{Ablation study on the impact of sampler $\pi_{\text{sam}}$:}
We conduct an ablation study to better understand the impact of samplers. 
We use iterative \texttt{DPO} as the base algorithm and consider two sampling subroutines:
\begin{enumerate}[leftmargin=*,topsep=0pt,itemsep=-1ex,partopsep=1ex,parsep=1ex]
    \item both responses are sampled by the policy of the previous iteration, i.e., $x\sim \rho, (y^1, y^2)\sim \pi_{t}(\cdot|x)$;
    \item  one response is sampled from the previous iteration’s policy and one from the initial policy, i.e., $x\sim \rho, y_1\sim \pi_t(\cdot|x),  y^2\sim \pi_{\text{ref}}(\cdot|x)$.
\end{enumerate}
As shown in Table~\ref{tab:2}, we study two metrics: 1). the reward corresponding to the responses produced by the models, 2). the win rate with respect to the base model $\pi_{\text{ref}}$. 
Notice that for both iteration 2 and iteration 3, the difference in win rate between the two sampler settings is relatively small, whereas the discrepancy in average reward is substantial.
In addition, we plot the reward distribution of the model outputs, as illustrated in Figures~\ref{fig:img1}. 
For samplers $(\pi_t, \pi_{\text{ref}})$, the reward distribution remains relatively unchanged between iteration 2 and 3.
In contrast, samplers $(\pi_t, \pi_t)$ demonstrates a more pronounced change in the reward distribution.
These results are consistent with our theoretical intuition in Section \ref{section:1}: collecting data by $(\pi_t, \pi_{\text{ref}})$ can result in $\pi_t$ consistently winning, thereby limiting its capacity to acquire new information.
Consequently, the models can only learn a policy that is sufficiently better than $\pi_{\text{ref}}$ (with 86\% and 89\% win rate), but fail to improve any further.

\textbf{Ablation study on the choices of $\beta$:}
Lastly, we conduct an ablation study to investigate the discrepancy between theoretical and empirical choices of the KL coefficient $\beta$.
According to Theorem~\ref{theo:SE-POPO}, a smaller $\beta$ is theoretically preferable, as regularization drives the policy away from optimality.
To examine this, we consider three choices of $\beta$: $\{0.1, 0.03, 0.01\}$.
Specifically, we adjust the exploration coefficient $\alpha$ in accordance with $\beta$ to keep $\alpha\beta$ constant, thereby ensuring that the scale of the exploration term's gradient remains stable.
The results, summarized in Table \ref{tab:beta} in Appendix~\ref{appendix-exp}, reveal that $\beta=0.03$ performs best, followed by $\beta=0.1$, and lastly $\beta=0.01$.
This suggests that while smaller $\beta$ values are theoretically desirable, an excessively small $\beta$ can introduce instability during training, leading to suboptimal performance in practice.


\section{Limitation \& Conclusion}
\label{sec:conclude}
In this work, we propose \texttt{SE-POPO}, the first practical and provably sample-efficient online exploration algorithm for RLHF with a polynomial dependence on the reward scale. 
\texttt{SE-POPO} offers a strictly superior sample complexity guarantee in theory, while outperforming existing baselines in practice.
One limitation of the approach is that \texttt{SE-POPO} does not extend to general preference models beyond Bradley–Terry model, particularly those where the preference is not necessarily monotonic.
Future directions include investigating online exploration algorithms with minimal length exploitation \citep{singhal2023long, meng2024simpo}, extending our algorithms to token-level MDP \citep{xie2024exploratory, zhong2024dpo} and multi-turn RLHF settings \citep{shani2024multi, gao2024regressing, xiong2024building}.

\bibliography{example_paper}
\bibliographystyle{plainnat}

\newpage
\appendix
\tableofcontents
\newpage

\section{Related Works}\label{sec:rw}

\paragraph{RLHF and RLHF algorithms}
The current RLHF framework was first
popularized by \citep{christiano2017deep}, which served to direct the attention of the deep RL community to the preference-based feedback.
Due to its significant success in LLM alignment \citep{chatgpt, touvron2023llama}, RLHF has gained substantial interest and become one of the prominent research topics in recent years.
The most widely adopted and standard RLHF framework, as described in \citep{ouyang2022training, touvron2023llama}, consists of two primary stages: 1) optimizing a reward model using the preference dataset, and 2) refining the LLM policy using PPO \citep{schulman2017proximal} based on the optimized reward model.
While this RLHF framework has achieved tremendous success in the industry, its adoption by academic and open-source communities is challenging due to the essential limitations of PPO, such as issues with reproducibility \citep{choshen2019weaknesses},  hyperparameters sensitivity \citep{engstrom2020implementation}, and its significant computational resource requirements.
Inspired by the limitations of this two-stage approach, a new line of research focuses on single-stage algorithms, including SLiC \citep{zhao2023slic}, DPO \citep{rafailov2024direct}, and its variants, such as IPO \citep{azar2024general}, SPPO \citep{wu2024self},
VPO \citep{cen2024value}, XPO \citep{xie2024exploratory}, and SELM \citep{zhang2024self}. 
These algorithms bypass the reward modeling step and learn a policy by optimizing a designed loss function on the preference dataset directly. 
It is observed that such algorithms are much more stable than PPO and achieve impressive performance on public benchmarks \citep{tunstall2023zephyr, dubois2024alpacafarm, zheng2023judging}.

\paragraph{Theoretical Study on RLHF}
The earliest theoretical frameworks for RLHF trace back to the dueling bandits literature \citep{yue2012k, saha2018battle, bengs2021preference}, along with studies considering tabular RL with finite state space \citep{xu2020preference, novoseller2020dueling} and linear RL or general function approximation RL with infinite state space \citep{pacchiano2021dueling,chen2022human, wu2023making, zhan2023query, das2024active, wang2023rlhf}. 
Apart from the online setting, a substantial body of research focuses on offline RLHF \citep{zhu2023principled, zhan2023provable, ji2024self, liu2024provably}, which leverages predetermined offline datasets with appropriate coverage conditions over the state-action space and can be considered complementary to our work.
Although these studies offer sample complexity guarantees for RLHF, most algorithms are not scalable enough to be applicable to modern LLMs with large transformer architectures.
For instance, \citet{pacchiano2021dueling, das2024active} incorporate exploration bonuses tailored for linear models in the reward estimation.
\citet{ chen2022human,zhan2023provable, wang2023rlhf} rely on model-based function approximation and explicitly estimate the policy confidence set. These approaches fail to yield efficient or practical algorithms when applied to LLMs.

\paragraph{Exploration for online LLM alignment}
Exploration in online RLHF has seen rapid development recently.
Earlier attempts, such as online DPO \citep{guo2024direct} and iterative DPO \citep{xu2023some, dong2024rlhf,xiong2024building}, primarily rely on passive exploration, i.e. the inherent randomness of LLM policy, and lack explicit mechanisms to encourage diverse and exploratory responses.
The importance of active exploration in RLHF has been highlighted by \citet{dwaracherla2024efficient}. 
Subsequently, \citet{ye2024theoretical, xiong2024iterative} propose algorithms with an active exploration mechanism and provide a sample complexity guarantees for online RLHF.
However, these exploration strategies involve solving an intractable optimization problem, making them impractical to implement in LLM alignment. 
Notably, in these works, experiments are often conducted based on heuristic variants of the proposed algorithms, resulting in a significant gap between theory and practice.
More recently, \citet{cen2024value,xie2024exploratory,zhang2024self} introduce implementation-friendly and provably sample-efficient exploration algorithms for RLHF, which are most relevant to our work.
All three papers are based on the common idea of augmenting the DPO loss with a \textit{reward-based} optimistic bonus to encourage exploration. 
Among them, \citet{zhang2024self, cen2024value} mainly focus on the exploration under the contextual bandit formulation of RLHF, whereas \citet{xie2024exploratory} provides analysis for the token-level MDP formulation.
However, a significant limitation of these algorithms is that their sample complexity scales exponentially with $ R_{\text{max}}$, the scale of the reward function (see Asm. \ref{ass: BT}), which is highly inefficient in both theory and practice. Our algorithm becomes the first that remove such $\exp(R_{\text{max}})$ dependency.

\section{Lightweight Implementation of \texttt{SE-POPO}}
\label{sec:lightweight}
In this section, we demonstrate that the moving from \eqref{popo-obj-dpo} to \eqref{popo-obj-dpo-r} is virtually a ``free-lunch'' reduction, based on the following neat observation.
\begin{lemma}
    Define 
    $H(r, \pi)=\underset{ 
 \substack{x\sim \rho\\ y'\sim \pi_{\text{sam}  }(\cdot|x)}}{\mathbb{E}}\left[  \sigma\left( - \beta \log \frac{\pi(y'|x)}{\pi_{\text{ref}}(y'|x)}\right)  \right],
    $
    then for every $r\in \mathcal{R}$, we have 
    \begin{align*}
        |G(r, \pi(r))- H(r, \pi(r))|\le \frac{\beta}{2}\mathbb{E}_{x\sim \rho}[\mathbb{D}_{\text{KL}}(\pi^\star_r(\cdot|x)||\pi_{\text{ref}}(\cdot|x))],
    \end{align*}
     where $\pi^\star_r = \arg\max_\pi \mathbb{E}_{x\sim \rho, y\sim \pi(\cdot|x)}[r(x,y)]$.
\label{lemma:POPO-efficient}
\end{lemma}
Lemma~\ref{lemma:POPO-efficient} implies that the gap between using $G(r, \pi(r))$ and $H(r, \pi(r))$ scales with \( \beta \) and the KL divergence between $\pi_r^\star$ and $\pi_{\text{ref}}$.
In this case, given \( \beta = o(1/\sqrt{T}) \), replacing $G(r, \pi(r))$ with $H(r, \pi(r))$ in the optimization objective (\ref{popo-obj-dpo-theory}) still guarantees that Theorem~\ref{theo:POPO} essentially holds, i.e.,
\begin{theorem}
    By replacing $G(r, \pi(r))$ in the optimization objective (\ref{popo-obj-dpo-theory}) with $H(r, \pi(r))$, with probability $1-2\delta$,  \texttt{POPO} guarantees that $\text{Reg}_{\text{pref}}(\pi_{\text{sam}}, T)\le $
    \begin{align*} 
{\mathcal{O}}\left( 
\sqrt{dR_{\text{max}}T\log \frac{TR_{\text{max}}}{d}\log\frac{|\mathcal{R}|}{\delta}}  + d\exp(R_{\text{max}})\log \frac{TR_{\text{max}}}{d}\log \frac{|\mathcal{R}|}{\delta} + \beta T  C_{\text{KL}}'  \right),
\end{align*}
    where $C_{\text{KL}}' = \max_{r\in \mathcal{R}}\mathbb{E}_{x\sim \rho}\left[ 
\mathbb{D}_{\text{KL}}(\pi_r^\star(\cdot|x)||\pi_{\text{ref}}(\cdot|x))  \right]$.
\label{theo:POPO-r}
\end{theorem}
Theorem~\ref{theo:POPO-r} establishes a preference regret bound that is fundamentally consistent with Theorem~\ref{theo:POPO}, with the only difference being in the KL term.   
In particular, when \(\beta\) is sufficiently small, Theorem~\ref{theo:POPO-r} reduces to Theorem~\ref{theo:POPO} immediately.  
Therefore, assuming \(C_{\text{KL}}'\) is well-bounded, it follows that the reward regret in Theorem~\ref{theo:SE-POPO} remains valid with the new exploration bonus.

\section{Supporting Lemmas}
We now present several auxiliary lemmas that will be used in next section's proof.
\begin{lemma}(MLE estimation error \citep{cen2024value, xie2024exploratory})
\label{lemma_MLE}
With probability at least $1-\delta$, for all $r\in \mathcal{R}$ and $t\in [T]$, there is 
\begin{align*}
    &\ell(r^\star, \mathcal{D}_{t-1})- \ell(r, \mathcal{D}_{t-1})\\
    &\le -\frac{1}{2}\sum_{s=1}^{t-1} \mathbb{E}_{x\sim \rho, (y,y')\sim \pi_s\otimes \pi_{\text{sam}}(\cdot|x)}\left[\left(\mathbb{P}^\star(y\succ y'|x)- \mathbb{P}_{r}(y\succ y'|x)\right)^2\right]+ 2\log \frac{|\mathcal{R}|}{\delta}.
\end{align*}
\end{lemma}

\begin{lemma}
\label{lemma:confidence:1}
    Define
    \begin{align*}
        \mathcal{R}(\mathcal{D}) = \left\{r\in \mathcal{R} \mid \ell(r, \mathcal{D})-\min_{r'\in \mathcal{R}}\ell(r', \mathcal{D})\le 2\log \frac{|\mathcal{R}|}{\delta}\right\}.
    \end{align*}
    Conditioning on Lemma~\ref{lemma_MLE}, for all $t\in [T]$, there is $r^\star\in \mathcal{R}(\mathcal{D}_t)$.
\end{lemma}
\begin{lemma}
\label{lemma:confidence:2}
    Conditioning on Lemma~\ref{lemma:confidence:1}, for all $t\in [T]$, there is 
    \begin{align*}
            r_{t+1} = \arg\max_{ r\in \mathcal{R}(\mathcal{D}_t)} \left\{ -\ell(r, \mathcal{D}_t) + \alpha G(r, \pi(r) \right\}
    \end{align*}
\end{lemma}
\begin{lemma}
\label{lemma:confidence:3}
With probability at least $1-\delta$, for all $r\in \mathcal{R}$ and $t\in [T]$, there is
\begin{align*}
    \sum_{s=1}^{t-1}\big(\mathbb{P}^\star(y_s\succ y_s'\mid x_s)&-\mathbb{P}_r(y_s\succ y_s'\mid x_s)\big)^2\\
    &\le 2\sum_{s=1}^{t-1}\mathbb{E}_{x\sim \rho, (y,y')\sim \pi_s\otimes \pi_{\text{sam}}(\cdot|x)}\left[\left(\mathbb{P}^\star(y\succ y'\mid x)-\mathbb{P}_r(y\succ y'\mid x)\right)^2\right] + \log\frac{|\mathcal{R}|}{\delta}.
\end{align*}
\end{lemma}
\section{Proof of Theorem~\ref{theo:POPO}}
By the definition of $G$, there is
\begin{align*}
    \text{Reg}_{\text{pref}}(\pi_{\text{sam}}, T)&\le \sum_{t=1}^T [G(r^\star, \pi^\star) - G(r^\star, \pi_t)]
    \\&=  \underbrace{\sum_{t=1}^T [G(r^\star, \pi_\beta^\star) - G(r_t, \pi_t)]}_{\textsc{Term 1}} + \underbrace{\sum_{t=1}^T [ G(r_t, \pi_t)- G(r^\star, \pi_t)]}_{\textsc{Term 2}} + \underbrace{\sum_{t=1}^T [ G(r^\star, \pi^\star)- G(r^\star, \pi_\beta^\star)]}_{\textsc{Term 3}}
\end{align*}
where $\pi_\beta^\star = \arg\max_{\pi\in \Pi} J(r^\star, \pi)$ and $r_t$ represents the reward corresponding to $\pi_t$, i.e., $\pi_t = \arg\max_{\pi\in \Pi} J(r_t, \pi)$.

\paragraph{Bounding \textsc{Term 1}}
Notice that in objective (\ref{popo-obj}), $\pi_t$ is completely dependent on $r_t$.
In this regard, the function $G$ can be considered as a function that depends only on the reward.
By Lemma~\ref{lemma:confidence:1} and \ref{lemma:confidence:2}, we have
\begin{align*}
    -\ell(r^\star, \mathcal{D}_{t-1})+\alpha G(r^\star, \pi_\beta^\star)\le -\ell(r_t, \mathcal{D}_{t-1})+\alpha G(r_t, \pi_t),
\end{align*}
thus
\begin{align*}
     G(r^\star, \pi_\beta^\star) - G(r_t, \pi_t)\le \frac{1}{\alpha}[\ell(r^\star, \mathcal{D}_{t-1})- \ell(r_t, \mathcal{D}_{t-1})].
\end{align*}
By Lemma~\ref{lemma_MLE}, it holds that 
\begin{align*}
        \textsc{Term 1}\le -\frac{1}{2 \alpha}\sum_{t=1}^T\sum_{s=1}^{t-1}\mathbb{E}_{x\sim \rho, (y,y')\sim \pi_s\otimes \pi_{\text{sam}}(\cdot|x)}\left[\left(\mathbb{P}^\star(y\succ y'|x)- \mathbb{P}_{r_t}(y\succ y'|x)\right)^2\right]+ \frac{2}{\alpha} T\log \frac{|\mathcal{R}|}{\delta}.
\end{align*}

\paragraph{Bounding \textsc{Term 2}}
By Assumption~\ref{assumption:2}, we can rewrite \textsc{Term 2} into 
\begin{align*}
    &\textsc{Term 2} \\
    &=\sum_{t=1}^T \mathbb{E}_{x\sim \rho, (y,y')\sim \pi_t\otimes \pi_{\text{sam}}(\cdot|x)}\left[\mathbb{P}_{r_t}\left(y\succ y'|x\right)-\mathbb{P}^\star\left(y\succ y'|x\right)  \right]\\
    &= \sum_{t=1}^T \mathbb{E}_{x\sim \rho, (y,y')\sim \pi_t\otimes \pi_{\text{sam}}(\cdot|x)}\left[\sigma \left(r_t(x,y) - r_t(x,y')\right)-\sigma\left(r^\star(x,y) - r^\star(x,y')\right)  \right]\\
    &\le \sum_{t=1}^T \mathbb{E}_{x\sim \rho, (y,y')\sim \pi_t\otimes \pi_{\text{sam}}(\cdot|x)}\left[\sigma \left(\langle\theta_t, \phi(x,y, y') \rangle\right)-\sigma\left(\langle\theta^\star, \phi(x,y, y') \rangle\right)  \right]\\
    &\le  \sum_{t=1}^T \mathbb{E}_{x\sim \rho, (y,y')\sim \pi_t\otimes \pi_{\text{sam}}(\cdot|x)}\left[ 
\min\left\{\max_{v\in [-|\theta^\star, \phi(x,y, y')|, |\theta_t, \phi(x,y, y')| ]}\dot{\sigma}(v)| \langle \theta_t-\theta^\star, \phi(x, y, y')  \rangle  |, 1 \right\} \right]\\
    &\le  \sum_{t=1}^T \mathbb{E}_{x\sim \rho, (y,y')\sim \pi_t\otimes \pi_{\text{sam}}(\cdot|x)}\left[ 
\min\left\{\exp(|\langle \theta_t-\theta^\star, \phi(x, y, y')  \rangle|)\dot{\sigma}(|\langle\theta^\star, \phi(x,y, y')\rangle|)| \langle \theta_t-\theta^\star, \phi(x, y, y')  \rangle  |, 1 \right\} \right]\\
    & \le \underbrace{ \sum_{t=1}^T \mathbb{E}_{x\sim \rho, (y,y')\sim \pi_t\otimes \pi_{\text{sam}}(\cdot|x)}\left[ 
\min\left\{|\langle \theta_t-\theta^\star, \phi(x, y, y')  \rangle|^2, 1 \right\}  \right]}_{\textsc{Term 2(1)}} \\
&\qquad \qquad \qquad+ 3 \underbrace{\sum_{t=1}^T \mathbb{E}_{x\sim \rho, (y,y')\sim \pi_t\otimes \pi_{\text{sam}}(\cdot|x)}\left[ \dot{\sigma}(|\langle\theta^\star, \phi(x,y, y')\rangle|)| \langle \theta_t-\theta^\star, \phi(x, y, y')  \rangle  |\right]}_{\textsc{Term 2(2)}}.
\end{align*}
Here, $\dot{\sigma}$ represents the derivative of the sigmoid function.
The last inequality is due to $\min\{\exp(a)b, 1\}\le \min\{a^2,1\}+ \exp(1)b$ for any $a,b\ge 0$.
Denote by 
\begin{align*}
    &W_t = \theta_{t}-\theta^\star,\ X_t = \phi(x_{t}, y_{t}, y_{t}'),\ w_t= \dot{\sigma}(|\langle\theta^\star, \phi(x_t,y_t, y_t')\rangle|), \\
    &Y_t = w_tX_t,\ \Lambda_t = \epsilon \mathbf{I}+ \sum_{s=1}^{t-1}X_sX_s^\top, \ \Sigma_t = \epsilon \mathbf{I}+ \sum_{s=1}^{t-1}Y_sY_s^\top,
\end{align*}
for some $\epsilon>0$.
We first focus on bounding \textsc{Term 2(1)}.
By the definition of $\theta_t$, it suffices to note that
\begin{align*}
    \|W_t\|^2_{\Lambda_t} &= {\epsilon \|W_t\|^2+ \sum_{s=1}^{t-1}\langle W_t, X_s \rangle^2 }\\
    &\le  {\epsilon R^2_{\text{max}}+ \sum_{s=1}^{t-1} ((r_t(x_s, y_s) - r_t(x_s, y_s'))-(r^\star(x_s, y_s) - r^\star(x_s, y_s'))) ^2 }\\
    &\le   {\epsilon R^2_{\text{max}}+ (1+\exp(R_{\text{max}}))\sum_{s=1}^{t-1} \left(\mathbb{P}^\star(y_s\succ y_s'|x_s)-\mathbb{P}_{r_t}(y_s\succ y_s'|x_s)\right)^2 }.
\end{align*}
By Lemma~\ref{lemma_MLE}, \ref{lemma:confidence:2} and \ref{lemma:confidence:3}, there is
\begin{align*}
    &\sum_{s=1}^{t-1} \left(\mathbb{P}^\star(y_s\succ y_s'|x_s)-\mathbb{P}_{r_t}(y_s\succ y_s'|x_s)\right)^2\\
    &\le 2\sum_{s=1}^{t-1}\mathbb{E}_{x\sim \rho, (y,y')\sim \pi_s\otimes \pi_{\text{sam}}(\cdot|x)}\left[\left(\mathbb{P}^\star(y\succ y'\mid x)-\mathbb{P}_{r_t}(y\succ y'\mid x)\right)^2\right] + \log\frac{|\mathcal{R}|}{\delta}\\
    &\le 9\log \frac{|\mathcal{R}|}{\delta}+ 4\ell(r_t,\mathcal{D}_{t-1})-4\ell(r^\star,\mathcal{D}_{t-1})\\
    &\le 9\log \frac{|\mathcal{R}|}{\delta}+ 4\ell(r_t,\mathcal{D}_{t-1})-4\min_{r'\in \mathcal{R}}\ell(r',\mathcal{D}_{t-1})\le17\log \frac{|\mathcal{R}|}{\delta}.
\end{align*}
Thus we have 
\begin{align*}
    \|W_t\|^2_{\Lambda_t}   \le  {\epsilon R^2_{\text{max}}+ 17(1+\exp(R_{\text{max}}))\log\frac{|\mathcal{R}|}{\delta} },\ \forall t,
\end{align*}
which means
\begin{align*}
    &\textsc{Term 2(1)} \\
    &\le \sum_{t=1}^T \mathbb{E}_{x\sim \rho, (y,y')\sim \pi_t\otimes \pi_{\text{sam}}(\cdot|x)}\left[ 
\min\left\{ \|W_t\|^2_{\Lambda_t}\| X_t\|^2_{\Lambda_t^{-1}}, 1 \right\}  \right]\\&\le \left({\epsilon R^2_{\text{max}}+ 4\exp(R_{\text{max}})\log\frac{|\mathcal{R}|T}{\delta} }\right)\left(\sum_{t=1}^T \mathbb{E}_{x\sim \rho, (y,y')\sim \pi_t\otimes \pi_{\text{sam}}(\cdot|x)}\left[ 
\min\left\{ \|X_t\|^2_{\Lambda_t^{-1}}, 1 \right\}  \right]\right).
\end{align*}
To proceed, we recall the elliptical potential lemma.
\begin{lemma}(\citep{abbasi2011improved}, Lemma 11)
\label{lemma_abbasi2011improved}
    Let $\{X_t\}$ be a sequence in $\mathbb{R}^d$ and $\Lambda_0\in \mathbb{R}^{d\times d}$ a positive definite matrix. Define $\Lambda_t = \Lambda_0+\sum_{s=1}^{t-1}X_sX_s^\top $, if $\|X_t\|_2\le L$ for all $t$ , there is 
    \begin{align*}
        \sum_{t=1}^T \min\left\{ 1, \|X_t\|_{\Lambda_{t}^{-1}}^2  \right\}\le 2(d\log(\text{trace}(\Lambda_0)+TL^2/d)-\log \det(\Lambda_0)).
    \end{align*}
\end{lemma}
Applying this lemma we can get 
\begin{align*}
    \sum_{t=1}^T \min\left\{ 1, \|X_t\|_{\Lambda_{t}^{-1}}^2  \right\}\le 2d\log \left(1+\frac{4TR^2_{\text{max}}/d}{\epsilon}\right):=d(\epsilon).
\end{align*}
Thus
\begin{align*}
    \sum_{t=1}^T \mathbb{E}_{x\sim \rho, (y,y')\sim \pi_t\otimes \pi_{\text{sam}}(\cdot|x)}\left[ 
\min\left\{ \|X_t\|_{\Lambda_t^{-1}}, 1 \right\}  \right] &=  \mathbb{E}_{\{x_t, y_t, y_t'\}_{t=1}^T}\left[ 
\sum_{t=1}^T\min\left\{ \|X_t\|^2_{\Lambda_t^{-1}}, 1 \right\}  \right]\le d(\epsilon)
\end{align*}
and 
\begin{align*}
    \textsc{Term 2(1)} \le {\epsilon d(\epsilon) R^2_{\text{max}}+ 17d(\epsilon)(1+\exp(R_{\text{max}}))\log\frac{|\mathcal{R}|}{\delta} }
\end{align*}
Now we start to bound \textsc{Term 2(2)}.
We decompose the term into
\begin{align}
    \textsc{Term 2(2)} &=  \mathbb{E}_{\{x_t, y_t, y_t'\}_{t=1}^T}\left[ \sum_{t=1}^T   | \langle W_t, Y_t  \rangle  |\right] \nonumber\\
    &= \mathbb{E}_{\{x_t, y_t, y_t'\}_{t=1}^T}\left[ \sum_{t=1}^T   | \langle W_t, Y_t  \rangle  |\mathbbm{1}\big\{\|Y_t\|_{\Sigma_t^{-1}}\le 1\big\}\right]\nonumber\\
    &\qquad\qquad\qquad+ \mathbb{E}_{\{x_t, y_t, y_t'\}_{t=1}^T}\left[ \sum_{t=1}^T   | \langle W_t, Y_t \rangle  |\mathbbm{1}\big\{\|Y_t\|_{\Sigma_t^{-1}}> 1\big\}\right].
    \label{appendix_term2_decompose}
\end{align}
Now we control the term terms in (\ref{appendix_term2_decompose}) respectively.
\begin{itemize}
    \item The first term is bounded by
    \begin{align*}
        &\sum_{t=1}^T |\langle W_t,  Y_t \rangle|\mathbbm{1}\big\{\|Y_t\|_{\Sigma_t^{-1}}\le 1\big\}\\
        &\le \sum_{t=1}^T \|W_t\|_{\Sigma_t}\|Y_t\|_{\Sigma_t^{-1}}\mathbbm{1}\big\{\|Y_t\|_{\Sigma_t^{-1}}\le 1\big\}\\
        &\le \sum_{t=1}^T \|W_t\|_{\Sigma_t} 
 \min\left\{1, \|Y_t\|_{\Sigma_t^{-1}}\right\}\\
 &= \sum_{t=1}^T \left[\epsilon \|W_t\|^2+ \sum_{s=1}^{t-1} \langle W_t, Y_s \rangle^2 \right]^{1/2}\bigg[\min\left\{1, \|Y_t\|^2_{\Sigma_t^{-1}}\right\}\bigg]^{1/2}\\
 &\le \left\{\sum_{t=1}^T\left[\epsilon \|W_t\|^2+ \sum_{s=1}^{t-1} \langle W_t, Y_s \rangle^2 \right]\right\}^{1/2}\bigg\{\sum_{t=1}^T \min\left\{1, \|Y_t\|^2_{\Sigma_t^{-1}}\right\}\bigg\}^{1/2}\\
 &\le \sqrt{d(\epsilon)\epsilon T R_{\text{max}}^2}+ \sqrt{d(\epsilon)} \left\{\sum_{t=1}^T\sum_{s=1}^{t-1} \langle W_t, Y_s \rangle^2 \right\}^{1/2}\\
 &\le \sqrt{d(\epsilon)\epsilon T R_{\text{max}}^2} + \frac{d(\epsilon)}{2\mu}+ \frac{\mu}{2}\sum_{t=1}^T\sum_{s=1}^{t-1} \langle W_t, Y_s \rangle^2,
    \end{align*}
    where the third inequality is due to Cauchy–Schwarz inequality, the fourth inequality is because $\sqrt{a+b}\le \sqrt{a}+\sqrt{b}$, and the last inequality is by Young’s inequality.
\item The second term is bounded by applying Lemma~\ref{lemma_abbasi2011improved}, i.e.,
\begin{align*}
    \sum_{t=1}^T |\langle W_t,  Y_t \rangle|\mathbbm{1}\big\{\|Y_t\|_{\Sigma_t^{-1}}> 1\big\}&\le R_{\text{max}}\sum_{t=1}^T \mathbbm{1}\big\{\|Y_t\|_{\Sigma_t^{-1}}> 1\big\}\\
    &\le R_{\text{max}}\sum_{t=1}^T  \min\left\{1, \|Y_i\|_{\Sigma_i^{-1}}\right\}\le R_{\text{max}} d(\epsilon).
\end{align*}
\end{itemize}
Summing up the two terms we arrive at
\begin{align*}
    \sum_{t=1}^T |\langle W_t,  Y_t \rangle|\le R_{\text{max}} d(\epsilon)+\sqrt{d(\epsilon)\epsilon T R_{\text{max}}^2} + \frac{d(\epsilon)}{2\mu}+ \frac{\mu}{2}\sum_{t=1}^T\sum_{s=1}^{t-1} \langle W_t, Y_s \rangle^2.
\end{align*}
thus
\begin{align*}
    \textsc{Term 2(2)} &\le R_{\text{max}} d(\epsilon)+\sqrt{d(\epsilon)\epsilon T R_{\text{max}}^2} + \frac{d(\epsilon)}{2\mu}+ \frac{\mu}{2} \mathbb{E}_{\{x_t, y_t, y_t'\}_{t=1}^T}\left[\sum_{t=1}^T \sum_{s=1}^{t-1} \langle W_t, Y_s \rangle^2 \right]
\end{align*}
As expectation of sum is sum of expectation, we have 
\begin{align*}
    &\mathbb{E}_{\{x_t, y_t, y_t'\}_{t=1}^T}\left[\sum_{t=1}^T \sum_{s=1}^{t-1} \langle W_t, Y_s \rangle^2 \right] = \sum_{t=1}^T\sum_{s=1}^{t-1}\mathbb{E}_{x_s\sim \rho, (y_s,y_s')\sim \pi_s\otimes \pi_{\text{sam}}(\cdot|x)}\left[ \langle W_t, Y_s \rangle^2  \right]\\
    &=\sum_{t=1}^T\sum_{s=1}^{t-1}\mathbb{E}_{x_s\sim \rho, (y_s,y_s')\sim \pi_s\otimes \pi_{\text{sam}}(\cdot|x)}\left[ \dot{\sigma}(\langle\theta^\star, X_s\rangle)^2 \left(\langle\theta_t, X_s\rangle - \langle\theta^\star, X_s\rangle\right)^2 \right].
\end{align*}
To proceed, we introduce an auxiliary lemma.
\begin{lemma}
\label{lemma_sigmopidlinkfunc}
    For any $a, b \in [-R_{\text{max}}/2, R_{\text{max}}/2]$, there is 
    \begin{align*}
        \dot{\sigma}(a)|a-b|\le 3 R_{\text{max}}|\sigma(a)- \sigma(b)|.
    \end{align*}
\end{lemma}
By Lemma~\ref{lemma_sigmopidlinkfunc}, we have 
\begin{align*}
    &\sum_{t=1}^T\sum_{s=1}^{t-1}\mathbb{E}_{x_s\sim \rho, (y_s,y_s')\sim \pi_s\otimes \pi_{\text{sam}}(\cdot|x)}\left[ \dot{\sigma}(\langle\theta^\star, X_s\rangle)^2 \left(\langle\theta_t, X_s\rangle - \langle\theta^\star, X_s\rangle\right)^2 \right]\\
    &\le 3R_{\text{max}} \sum_{t=1}^T\sum_{s=1}^{t-1}\mathbb{E}_{x_s\sim \rho, y_s\sim \pi_s(\cdot|x), y_s'\sim \pi_{\text{sam}}(\cdot|x)}\left[ \left({\sigma}(\langle\theta_t, X_s\rangle)-{\sigma}(\langle\theta^\star, X_s\rangle)\right)^2 \right]\\
    &= 3R_{\text{max}} \sum_{t=1}^T\sum_{s=1}^{t-1}\mathbb{E}_{x\sim \rho, y\sim \pi_s(\cdot|x), y'\sim \pi_{\text{sam}}(\cdot|x)}\left[ \left(\mathbb{P}_{r_t}(y\succ y'|x) - \mathbb{P}^\star(y\succ y'|x) \right)^2 \right].
\end{align*}
Combining the above, we finally get
\begin{align*}
    \textsc{Term 2}&\le R_{\text{max}} d(\epsilon)+\sqrt{d(\epsilon)\epsilon T R_{\text{max}}^2} + \frac{d(\epsilon)}{2\mu}+ {\epsilon d(\epsilon) R^2_{\text{max}}+ 17d(\epsilon)(1+\exp(R_{\text{max}}))\log\frac{|\mathcal{R}|}{\delta} }\\
    &+ \frac{3\mu R_{\text{max}}}{2}\sum_{t=1}^T\sum_{s=1}^{t-1} \mathbb{E}_{x\sim \rho, y\sim \pi_s(\cdot|x), y'\sim \pi_{\text{sam}}(\cdot|x)}\left[ \left(\mathbb{P}_{r_t}\left(y\succ y'|x\right)-\mathbb{P}^\star\left(y\succ y'|x\right) \right) ^2 \right].
\end{align*}
\paragraph{Bounding \textsc{Term 3}}
By the choice of $\pi_t$ in (\ref{popo-obj}), we have $ J(r^\star, \pi^\star)\le J(r^\star, \pi^\star_\beta)$.
This implies that
\begin{align*}
    \mathbb{E}_{x\sim \rho, (y^\star, y)\sim \pi^\star\otimes \pi_\beta^\star(\cdot|x)}\left[ r^\star(x,y^\star) - r^\star(x,y) \right]\le \mathbb{E}_{x\sim \rho, (y^\star, y)\sim \pi^\star\otimes \pi_\beta^\star(\cdot|x)}\left[\beta\log \frac{\pi^\star(y^\star|x)}{\pi_{\text{ref}}(y^\star|x)}  -\beta\log \frac{\pi_\beta^\star(y|x)}{\pi_{\text{ref}}(y|x)} \right].
\end{align*}
The key observation is that for any $y'\in \mathcal{Y}$, there is
\begin{align*}
    r^\star(x,y^\star) - r^\star(x,y) \ge 4[\mathbb{P}^\star(y^\star\succ y'|x)- \mathbb{P}^\star(y\succ y'|x)].
\end{align*}
This is because $y^\star$ is always the best response, which means that $r^\star(x,y^\star) \ge r^\star(x,y)$ for sure. 
Moreover, the gradient of sigmoid function is less than $1/4$, thereby the gap between the preferences is at most $1/4$th of the gap between rewards. 
Using the inequality, we have
\begin{align*}
     &\mathbb{E}_{x\sim \rho, (y^\star, y,y')\sim \pi^\star\otimes \pi_\beta^\star\otimes \pi_{\text{sam}}(\cdot|x)}\big[\mathbb{P}^\star(y^\star\succ y'|x)- \mathbb{P}^\star(y\succ y'|x) \big]\\
     &\le \frac{1}{4} \mathbb{E}_{x\sim \rho, (y^\star, y)\sim \pi^\star\otimes \pi_\beta^\star(\cdot|x)}\left[\beta\log \frac{\pi^\star(y^\star|x)}{\pi_{\text{ref}}(y^\star|x)}  -\beta\log \frac{\pi_\beta^\star(y|x)}{\pi_{\text{ref}}(y|x)} \right]\\
     &\le \frac{1}{4}\mathbb{E}_{x\sim \rho, y^\star\sim \pi^\star(\cdot|x)}\left[\beta\log \frac{\pi^\star(y^\star|x)}{\pi_{\text{ref}}(y^\star|x)} \right]\\
     &= \frac{\beta}{4}\mathbb{E}_{x\sim \rho}\left[ \mathbb{D}_{\text{KL}}(\pi^\star(\cdot|x)||\pi_{\text{ref}}(\cdot|x))\right],
\end{align*}
Thus we have $\textsc{Term 3} \le \mathcal{O}(\beta T\mathbb{E}_{x\sim \rho}\left[ \mathbb{D}_{\text{KL}}(\pi^\star(\cdot|x)||\pi_{\text{ref}}(\cdot|x))\right])$.

\paragraph{Finishing up}
Combining \texttt{Term 1} and \texttt{Term 2}, with probability $1-2\delta$ and $\epsilon= \frac{1}{T}$, there is
\begin{align*}
     \sum_{t=1}^T [G(r^\star, \pi_\beta^\star) - G(r^\star, \pi_t)]\le {\mathcal{O}}\left(\frac{1}{\alpha}T \log \frac{|\mathcal{R}|}{\delta}+ \frac{d(\epsilon)}{\mu} + d(\epsilon)\exp(R_{\text{max}})\log \frac{|\mathcal{R}|}{\delta} \right).
\end{align*}
as long as $ \frac{3R_{\text{max}}\mu}{2}\le \frac{1}{2 \alpha}$.
Setting $\alpha = \sqrt{\frac{d\log \frac{T}{d}}{R_{\text{max}}T\log \frac{|\mathcal{R}|}{\delta} }}, \mu = \frac{1}{3}\sqrt{\frac{T\log \frac{|\mathcal{R}|}{\delta} }{R_{\text{max}}d\log \frac{T}{d}}}$, we finally bound $\text{Reg}_{\text{pref}}(\pi_{\text{sam}}, T)$ by
\begin{align*} 
 {\mathcal{O}}\left( 
\sqrt{dR_{\text{max}}T\log \frac{TR_{\text{max}}}{d}\log\frac{|\mathcal{R}|}{\delta}}  + d\exp(R_{\text{max}})\log \frac{TR_{\text{max}}}{d}\log \frac{|\mathcal{R}|}{\delta} + \beta T  C_{\text{KL}}  \right),
\end{align*}
Therefore, for $T\ge \tilde{\mathcal{O}}({d\exp(2R_{\text{max}})\log \frac{|\mathcal{R}|}{\delta}}/{R_{\text{max}}} )$, we have
\begin{align*}
    \underset{ 
 \substack{x\sim \rho\\  (y^\star, y, y')\sim \pi^\star\otimes\bar \pi \otimes \pi_{\text{sam}}(\cdot|x)}}{\mathbb{E}}\bigg[\mathbb{P}^\star(y^\star\succ y'|x)-\mathbb{P}^\star(y\succ y'|x)  \bigg]&= \frac{1}{T}\text{Reg}_{\text{pref}}(\pi_{\text{sam}}, T)\\
 &\le \tilde{{\mathcal{O}}}\left( \sqrt{\frac{{dR_{\text{max}}\log\frac{|\mathcal{R}|}{\delta}}}{T}} + \beta  C_{\text{KL}}  \right),
\end{align*}
which completes the proof.

\section{Proof of Theorem~\ref{theo:SE-POPO}}
For every $k=1,\dots, K$, by Theorem~\ref{theo:POPO}, with probability $1-\delta$, there is 
\begin{align}
        &\mathbb{E}_{x\sim \rho, (y^\star,y,y') \sim \pi^\star\otimes \pi_{\text{sam}}^{k+1} \otimes \pi_{\text{sam}}^k (\cdot|x)}\bigg[\mathbb{P}^\star(y^\star\succ y'|x)-P^\star(y\succ y'|x)  \bigg] \nonumber \\
        &= \frac{\text{Reg}_{\text{pref}}(\pi_{\text{sam}},T)}{T}
        \le \tilde{{\mathcal{O}}}\left( \sqrt{\frac{{dR_{\text{max}}\log\frac{|\mathcal{R}|}{\delta}}}{T}} + \beta  C_{\text{KL}}  \right)
\label{appendix-proof-of-theorem3.3-1}
    \end{align}
By Lemma~\ref{Pre2Rwd-reduction}, we have
\begin{align}
    &\mathbb{E}_{x\sim \rho, (y^\star,y,y') \sim \pi^\star\otimes \pi_{\text{sam}}^{k+1} \otimes \pi_{\text{sam}}^k (\cdot|x)}\bigg[\mathbbm{1}\{r^\star(x,y)-r^\star(x,y')\le 1\} \left[{r^\star(x,y^\star)-r^\star(x,y)}\right]  \bigg]\nonumber \\
    &
        \le \tilde{{\mathcal{O}}}\left( \sqrt{\frac{{dR_{\text{max}}^3\log\frac{|\mathcal{R}|}{\delta}}}{T}} + \beta  R_{\text{max}}C_{\text{KL}}  \right) \eqqcolon \text{Gap}(T)
\label{appendix-proof-of-theorem3.3-3}
\end{align}
For notation simplicity, we denote $r^\star(x,y)- r^\star(x,y')$ by $\Delta(x,y,y')$.
To proceed, we note that
\begin{align*}
\mathbbm{1}\big\{\Delta(x,y,y')\le 1\big\}\ge \mathbbm{1}\big\{ \Delta(x,y^\star,y)> \max(R_{\text{max}}-k,1)\big\}\mathbbm{1}\big\{ \Delta(x,y^\star,y')\le \max(R_{\text{max}}-k+1,1)\big\}.
\end{align*}
This is because when $\Delta(x,y^\star,y)> \max(R_{\text{max}}-k,1)$ and $\Delta(x,y^\star,y')\le \max(R_{\text{max}}-k+1,1)$, we have
\begin{align*}
    \Delta(x,y,y') &= \Delta(x,y^\star,y')-\Delta(x,y^\star,y)\\
    &\le \max(R_{\text{max}}-k+1,1)-\max(R_{\text{max}}-k,1)\le 1.
\end{align*}
In this regard, given $r^\star(x,y^\star) - r^\star(x,y)\ge 0$ for sure, we have
\begin{align*}
    &\mathbb{E}_{x\sim \rho, (y^\star,y,y') \sim \pi^\star\otimes \pi_{\text{sam}}^{k+1} \otimes \pi_{\text{sam}}^k (\cdot|x)}\bigg[\mathbbm{1}\{r^\star(x,y)-r^\star(x,y')\le 1\} \left[{r^\star(x,y^\star)-r^\star(x,y)}\right]  \bigg]\\
    &=\mathbb{E}_{x\sim \rho, (y^\star,y,y') \sim \pi^\star\otimes \pi_{\text{sam}}^{k+1} \otimes \pi_{\text{sam}}^k (\cdot|x)}\bigg[\mathbbm{1}\{\Delta(x,y,y')\le 1\} \Delta(x,y^\star,y)  \bigg]\\
    &\ge \mathbb{E}_{x\sim \rho, (y^\star,y,y') \sim \pi^\star\otimes \pi_{\text{sam}}^{k+1} \otimes \pi_{\text{sam}}^k (\cdot|x)}\bigg[ \mathbbm{1}\big\{ \Delta(x,y^\star,y)> \max(R_{\text{max}}-k,1)\big\}\\
    &\qquad \qquad \qquad \qquad  \qquad \qquad  \mathbbm{1}\big\{ \Delta(x,y^\star,y')\le \max(R_{\text{max}}-k+1,1)\big\}\Delta(x,y^\star,y)  \bigg]\\
    &\ge\mathbb{E}_{x\sim \rho, (y^\star,y,y') \sim \pi^\star\otimes \pi_{\text{sam}}^{k+1} \otimes \pi_{\text{sam}}^k (\cdot|x)}\bigg[ \mathbbm{1}\big\{ \Delta(x,y^\star,y)> \max(R_{\text{max}}-k,1)\big\}\\
    &\qquad \qquad \qquad \qquad \qquad \qquad  \qquad \qquad \ \  \mathbbm{1}\big\{ \Delta(x,y^\star,y')\le \max(R_{\text{max}}-k+1,1)\big\}  \bigg]\\
    &\ge\mathbb{E}_{x\sim \rho, (y^\star,y') \sim \pi^\star\otimes \pi_{\text{sam}}^k (\cdot|x)}\bigg[ \mathbbm{1}\big\{ \Delta(x,y^\star,y')\le \max(R_{\text{max}}-k+1,1)\big\}\bigg]\\
    &  \qquad \qquad \qquad \qquad \    - \mathbb{E}_{x\sim \rho, (y^\star,y) \sim \pi^\star\otimes \pi_{\text{sam}}^{k+1} }\bigg[ \mathbbm{1}\big\{ \Delta(x,y^\star,y)\le \max(R_{\text{max}}-k,1)\big\}\bigg].
\end{align*}
The second inequality is because the inner term is non-zero only if $\Delta(x,y^\star,y)> \max(R_{\text{max}}-k,1)\ge 1$.
Combining this with (\ref{appendix-proof-of-theorem3.3-3}), with probability $1-K\delta$, there is
\begin{align*}
    &\mathbb{E}_{x\sim \rho, (y^\star,y) \sim \pi^\star\otimes \pi_{\text{sam}}^{K+1}(\cdot|x) }\bigg[ \mathbbm{1}\big\{ \Delta(x,y^\star,y)\le \max(R_{\text{max}}-K,1)\big\}\bigg]\\
    &\ge \mathbb{E}_{x\sim \rho, (y^\star,y') \sim \pi^\star\otimes \pi_{\text{sam}}^K (\cdot|x)}\bigg[ \mathbbm{1}\big\{ \Delta(x,y^\star,y')\le \max(R_{\text{max}}-K+1,1)\big\}\bigg] - \text{Gap}(T)\\
    &\ge \mathbb{E}_{x\sim \rho, (y^\star,y') \sim \pi^\star\otimes \pi_{\text{sam}}^1 (\cdot|x)}\bigg[ \mathbbm{1}\big\{ \Delta(x,y^\star,y')\le \max(R_{\text{max}},1)\big\}\bigg] - K\text{Gap}(T)\\
    &= 1- \tilde{{\mathcal{O}}}\left( K\sqrt{\frac{{dR_{\text{max}}^3\log\frac{|\mathcal{R}|}{\delta}}}{T}} + K \beta  R_{\text{max}}C_{\text{KL}}  \right).
\end{align*}
Setting $K=\lceil  R_{\max} \rceil-1$, we achieve that
\begin{align*}
    \mathbb{E}_{x\sim \rho, (y^\star,y) \sim \pi^\star\otimes \pi_{\text{sam}}^{\lceil  R_{\max} \rceil} }&\bigg[ \mathbbm{1}\big\{ \Delta(x,y^\star,y)> 1\big\}\bigg]\\
    &\le  \tilde{{\mathcal{O}}}\left( \sqrt{\frac{{dR_{\text{max}}^5\log\frac{|\mathcal{R}|}{\delta}}}{T}} + \beta  R^2_{\text{max}}C_{\text{KL}}  \right).
\end{align*}
This result implies that
\begin{align*}
    \mathbb{E}_{x\sim \rho, (y,y') \sim \bar \pi\otimes \pi_{\text{sam}}^{\lceil  R_{\max} \rceil} }&\bigg[ \mathbbm{1}\big\{ \Delta(x,y,y')> 1\big\}\bigg]\\
    &\le  \tilde{{\mathcal{O}}}\left( \sqrt{\frac{{dR_{\text{max}}^5\log\frac{|\mathcal{R}|}{\delta}}}{T}} + \beta  R^2_{\text{max}}C_{\text{KL}}  \right).
\end{align*}
for all $\bar \pi$. 
In this regard, it suffices to note that $\pi_{\text{sam}}^{\lceil  R_{\max} \rceil}$ is a ``good enough'' sampler: it can return a response $y'$ such that $\Delta(x,y,y')\le 1$ with high probability.
Denote by $\bar \pi = \texttt{POPO}(\pi_{\text{ref}}, \pi_{\text{sam}}^{\lceil  R_{\max} \rceil}, T)$, with probability $1-\delta$, there is 
\begin{align*}
&\mathbb{E}_{x\sim\rho, (y^\star, y)\sim\pi^\star\otimes\bar \pi(\cdot|x)}\left[r^\star(x,y^\star)-r^\star(x,y)\right]\\
    &=\mathbb{E}_{x\sim\rho, (y^\star, y,y')\sim\pi^\star\otimes\bar \pi\otimes \pi_{\text{sam}}^{\lceil  R_{\max} \rceil}(\cdot|x)}\left[\mathbbm{1}\{\Delta(x,y,y')\le 1\}[r^\star(x,y^\star)-r^\star(x,y)]\right]\\
    &+\mathbb{E}_{x\sim\rho, (y^\star, y,y')\sim\pi^\star\otimes\bar \pi\otimes \pi_{\text{sam}}^{\lceil  R_{\max} \rceil}(\cdot|x)}\left[\mathbbm{1}\{\Delta(x,y,y')>1\}[r^\star(x,y^\star)-r^\star(x,y)]\right]\\
    &\le \mathbb{E}_{x\sim\rho, (y^\star, y,y')\sim\pi^\star\otimes\bar \pi\otimes \pi_{\text{sam}}^{\lceil  R_{\max} \rceil}(\cdot|x)}\left[\mathbbm{1}\{\Delta(x,y,y')\le 1\}[r^\star(x,y^\star)-r^\star(x,y)]\right]\\
    &+R_{\max}\mathbb{E}_{x\sim\rho, (y,y')\sim\bar \pi\otimes \pi_{\text{sam}}^{\lceil  R_{\max} \rceil}(\cdot|x)}\left[\mathbbm{1}\{\Delta(x,y,y')>1\}\right]\\
    &\le \tilde{{\mathcal{O}}}\left( \sqrt{\frac{{dR_{\text{max}}^7\log\frac{|\mathcal{R}|}{\delta}}}{T}} + \beta  R^3_{\text{max}}C_{\text{KL}}  \right).
\end{align*}
Setting $\beta\le o\left(\frac{1}{\sqrt{T}}\right), T= N/\lceil R_{\text{max}}\rceil$ and resizing $\delta = \delta/\lceil {R_{\max}} \rceil$ immediately complete the proof.

\section{Proof of Theorem~\ref{theo:POPO-r}}
In the proof of Theorem~\ref{theo:POPO}, the only place where we use the condition that \( \pi_{t+1} \) is the optimal solution to objective (\ref{popo-obj-dpo}) is in the proof of bounding \textsc{Term 1}. 
Therefore, it suffices to focus on \textsc{Term 1} itself.
As Lemma~\ref{lemma:confidence:1} and \ref{lemma:confidence:2} still hold, we have
\begin{align*}
    -\ell(r^\star, \mathcal{D}_{t-1})+\alpha H(r^\star, \pi_\beta^\star)\le -\ell(r_t, \mathcal{D}_{t-1})+\alpha H(r_t, \pi_t),
\end{align*}
Using Lemma~\ref{lemma:POPO-efficient}, it suffices to note
\begin{align*}
&-\ell(r^\star, \mathcal{D}_{t-1})+\alpha G(r^\star, \pi_\beta^\star)-\frac{\alpha\beta}{2}\max_{r\in \mathcal{R}}\mathbb{E}_{x\sim \rho}\left[ 
\mathbb{D}_{\text{KL}}(\pi_r^\star(\cdot|x)||\pi_{\text{ref}}(\cdot|x))  \right]\le
    -\ell(r^\star, \mathcal{D}_{t-1})+\alpha H(r^\star, \pi_\beta^\star)\\
    &-\ell(r_t, \mathcal{D}_{t-1})+\alpha H(r_t, \pi_t)\le -\ell(r_t, \mathcal{D}_{t-1})+\alpha G(r_t, \pi_t)+ \frac{\alpha\beta}{2}\max_{r\in \mathcal{R}}\mathbb{E}_{x\sim \rho}\left[ 
\mathbb{D}_{\text{KL}}(\pi_r^\star(\cdot|x)||\pi_{\text{ref}}(\cdot|x))  \right],
\end{align*}
thus
\begin{align*}
     G(r^\star, \pi_\beta^\star) - G(r_t, \pi_t)\le \frac{1}{\alpha}[\ell(r^\star, \mathcal{D}_{t-1})- \ell(r_t, \mathcal{D}_{t-1})]+\beta \max_{r\in \mathcal{R}}\mathbb{E}_{x\sim \rho}\left[ 
\mathbb{D}_{\text{KL}}(\pi_r^\star(\cdot|x)||\pi_{\text{ref}}(\cdot|x))  \right].
\end{align*}
This completes the proof.

\section{Proof of Lemma~\ref{Pre2Rwd-reduction}}
    Assuming $r^\star(x,y)\le r^\star(x,y')+1$.
    In this case, we note that 
    \begin{align*}
        P^\star(y\succ y'|x)= \frac{\exp(r^\star(x,y)-r^\star(x,y'))}{1+\exp(r^\star(x,y)-r^\star(x,y'))}\le \frac{e}{1+e}\le \frac{3}{4}.
    \end{align*}
    Given this, it suffices to focus on the case where $P^\star(y^\star \succ y'|x)\le 4/5$, otherwise
    \begin{align*}
        P^\star(y^\star\succ y'|x)-P^\star(y\succ y'|x)\ge \frac{4}{5}-\frac{3}{4} \ge \frac{r^\star(x,y^\star) - r^\star(x,y)}{20R_{max}}.
    \end{align*}
    Similarly, since $P^\star(y^\star\ge y'|x)\ge 1/2$, it suffices to focus on the case where $P^\star(y \succ y'|x)\ge 9/20$, otherwise
    \begin{align*}
        P^\star(y^\star\succ y'|x)-P^\star(y\succ y'|x)\ge \frac{1}{2}-\frac{9}{20} \ge \frac{r^\star(x,y^\star) - r^\star(x,y)}{20R_{max}}.
    \end{align*}
    In this way, we obtain certain constraints on the preferences $P^\star(y^\star\succ y'|x)$ and $P^\star(y\succ y'|x)$.
This further leads to constraints on the differences in rewards, i.e.,
\begin{align*}
   0 \le r^\star(x,y^\star) - r^\star(x,y')\le \frac{3}{2},\ -\frac{1}{2} \le r^\star(x,y) - r^\star(x,y')\le 1.
\end{align*}
Thus, it suffices to focus on the interval $[-\frac{1}{2}, \frac{3}{2}]$.
It is easily to see that
\begin{align*}
&P^\star(y^\star\succ y'|x)-P^\star(y\succ y'|x)\\
&=\sigma(r^\star(x,y^\star) - r^\star(x,y')) - \sigma(r^\star(x,y) - r^\star(x,y'))\\
    &\ge \min_{\Delta\in \left[-\frac{1}{2}, \frac{3}{2}\right]} \nabla \sigma(\Delta) [r^\star(x,y^\star) - r^\star(x,y')-(r^\star(x,y) - r^\star(x,y'))]\\
    & = \frac{r^\star(x,y^\star)-r^\star(x,y)}{20}.
\end{align*}
Combining the above we have 
\begin{align*}
    r^\star(x,y^\star)-r^\star(x,y)\le 20R_{\text{max}}[P^\star(y^\star\succ y'|x)-P^\star(y\succ y'|x)]
\end{align*}
with $r^\star(x,y)- r^\star(x,y')\le 1$.
This completes the proof.

\section{Proof of Lemma~\ref{lemma:POPO-efficient}}
Fix $r\in \mathcal{R}$.
Recall $\pi(r)$ is the optimal solution of the KL-regularized reward objective and $\pi^\star_r = \arg \max_{\pi} \mathbb{E}_{x\sim \rho, y\sim \pi(\cdot|x)}  [r(x,y)]$.
By the analysis of bounding \textsc{Term 3} in the proof of Theorem~\ref{theo:POPO}, we first note that 
\begin{align*}
    G(r, \pi_r^\star)- \frac{\beta}{4}\mathbb{E}_{x\sim \rho}[\mathbb{D}_{\text{KL}}(\pi^\star_r(\cdot|x)||\pi_{\text{ref}}(\cdot|x))]\le  G(r, \pi(r)) \le G(r, \pi_r^\star).
\end{align*}
It suffices to focus on $G(r, \pi_r^\star)$.
Then, we have
\begin{align*}
    G(r, \pi_r^\star) &= \mathbb{E}_{x\sim \rho, y\sim \pi_r^\star(\cdot|x), y'\sim \pi_{\text{sam}}(\cdot|x)}\left[\sigma(r(x,y)- r(x,y'))  \right]\\
    &=\mathbb{E}_{x\sim \rho, y\sim \pi_r^\star(\cdot|x), y'\sim \pi_{\text{sam}}(\cdot|x)}\left[\sigma\left(\beta \log \frac{\pi_r(y|x)}{\pi_{\text{ref}}(y|x)}- \beta \log \frac{\pi_r(y'|x)}{\pi_{\text{ref}}(y'|x)}\right)  \right]
\end{align*}
where $\pi_r=\pi(r)$.
Since here $y$ represents the response with the highest reward under $r$, it suffices to note that $\pi_r(y|x)\ge \pi_\text{ref}(y|x)$.
In this case, $\beta \log \frac{\pi_r(y|x)}{\pi_{\text{ref}}(y|x)}$ can be bounded by $[0, \beta\log \frac{1}{\pi_{\text{ref}}(y|x)}]$.
By the smoothness of sigmoid function, there is
\begin{align*}
&\mathbb{E}_{x\sim \rho, y'\sim \pi_{\text{sam}}(\cdot|x)}\left[\sigma\left(- \beta \log \frac{\pi_r(y'|x)}{\pi_{\text{ref}}(y'|x)}\right)  \right]\\
    &\le \mathbb{E}_{x\sim \rho, y\sim \pi_r^\star(\cdot|x), y'\sim \pi_{\text{sam}}(\cdot|x)}\left[\sigma\left(\beta \log \frac{\pi_r(y|x)}{\pi_{\text{ref}}(y|x)}- \beta \log \frac{\pi_r(y'|x)}{\pi_{\text{ref}}(y'|x)}\right)  \right]\\
    &\le\mathbb{E}_{x\sim \rho,y\sim \pi_r^\star(\cdot|x), y'\sim \pi_{\text{sam}}(\cdot|x)}\left[\sigma\left(- \beta \log \frac{\pi_r(y'|x)}{\pi_{\text{ref}}(y'|x)}\right)  +\frac{\beta}{4}\log \frac{\pi_r(y|x)}{\pi_{\text{ref}}(y|x)}\right]\\
    &\le \mathbb{E}_{x\sim \rho, y'\sim \pi_{\text{sam}}(\cdot|x)}\left[\sigma\left(- \beta \log \frac{\pi_r(y'|x)}{\pi_{\text{ref}}(y'|x)}\right) \right]+\frac{\beta}{4}\mathbb{E}_{x\sim \rho}[\mathbb{D}_{\text{KL}}(\pi^\star_r(\cdot|x)||\pi_{\text{ref}}(\cdot|x))]
\end{align*}
The last inequality is due to $q = \arg\max_p \sum_y q(y)\log p(y)$.
Combining the above we can conclude
\begin{align*}
    \left|G(r,\pi(r))-\mathbb{E}_{x\sim \rho, y'\sim \pi_{\text{sam}}(\cdot|x)}\left[\sigma\left(- \beta \log \frac{\pi_r(y'|x)}{\pi_{\text{ref}}(y'|x)}\right)  \right]\right|\le \frac{\beta}{2}\mathbb{E}_{x\sim \rho}[\mathbb{D}_{\text{KL}}(\pi^\star_r(\cdot|x)||\pi_{\text{ref}}(\cdot|x))].
\end{align*}
This completes the proof.

\section{Proof of Auxiliary Lemmas}
\subsection{Proof of Lemma~\ref{lemma_MLE}}
The proof refers to the proof of Lemma 2 in \citep{cen2024value}.
To begin with, there is 
\begin{align*}
    \ell(r^\star, \mathcal{D}_{t-1}) - \ell(r, \mathcal{D}_{t-1}) = -\sum_{s=1}^{t-1}\log \frac{\mathbb{P}_{r^\star}(y_s^+\succ y_s^-|x_s)}{\mathbb{P}_{r}(y_s^+\succ y_s^-|x_s)}.
\end{align*}
Define
\begin{align*}
    X_r^s = \log \frac{\mathbb{P}_{r^\star}(y_s^+\succ y_s^-|x_s)}{\mathbb{P}_{r}(y_s^+\succ y_s^-|x_s)}.
\end{align*}
Recall a martingale exponential inequality.
\begin{lemma} (\citep{zhang2023mathematical}, Theorem 13.2)
    Let $\{X_t\}_{t=1}^\infty$ be a sequence of random variables adapted to filtration $\{\mathcal{F}_t\}_{t=1}^\infty$. It holds with probability $1-\delta$ such that for any $t\ge 1$,
    \begin{align*}
        -\sum_{s=1}^t X_s \le \sum_{s=1}^t \log \mathbb{E}[\exp(-X_s)|\mathcal{F}_{s-1}] +\log \frac{1}{\delta}.
    \end{align*}
\end{lemma}
Notice that $\{X_r^t\}_{t=1}^\infty$ is a sequence of random variables adapted to filtration $\{\mathcal{F}_t\}_{t=1}^\infty$ with $\mathcal{F}_t$ given by the $\sigma$-algebra of $\{(x_s, y_s^+, y_s^-): s\le t\}$.
Applying the above lemma and taking a union bound among all $r\in \mathcal{R}$, we have with probability $1-\delta$, for every $r\in \mathcal{R}$ and $t$, there is
\begin{align*}
    -\frac{1}{2}\sum_{s=1}^{t-1} X_r^s &\le \sum_{s=1}^{t-1} \log \mathbb{E}\left[\exp\left(-\frac{1}{2}X_r^s\right)\bigg|\mathcal{F}_{s-1}\right] +\log \frac{|\mathcal{R}|}{\delta}\\
    &\le  \sum_{s=1}^{t-1}  \left(\mathbb{E}\left[ \exp\left(-\frac{1}{2}X_r^s\right)\bigg|\mathcal{F}_{s-1}\right]-1\right) +\log \frac{|\mathcal{R}|}{\delta},
\end{align*}
where the last inequality is due to $\log(1+x)\le x$ for all $x\ge -1$.
To proceed, note that
\begin{align*}
    \mathbb{E}\left[ \exp\left(-\frac{1}{2}X_r^s\right)\bigg|\mathcal{F}_{s-1}\right]
    &\le \mathbb{E}\left[ \sqrt{\frac{\mathbb{P}_{r}(y_s^+\succ y_s^-|x_s)}{\mathbb{P}_{r^\star}(y_s^+\succ y_s^-|x_s)}}  \bigg|\mathcal{F}_{s-1}\right]\\
    &= \mathbb{E}_{x\sim\rho, (y,y')\sim \pi_s\otimes \pi_{\text{sam}}(\cdot|x), (+,-)\sim \mathbb{P}_{r^\star}(\cdot|x,y,y')}\left[ \sqrt{\frac{\mathbb{P}_{r}(y^+\succ y^-|x)}{\mathbb{P}_{r^\star}(y^+\succ y^-|x)}} \right]\\
    & = \mathbb{E}_{x\sim\rho, (y,y')\sim \pi_s\otimes \pi_{\text{sam}}(\cdot|x)}\left[ \sum_{(+,-)}\sqrt{{\mathbb{P}_{r}(y^+\succ y^-|x)}{\mathbb{P}_{r^\star}(y^+\succ y^-|x)}} \right]\\
    & = 1- \frac{1}{2} \mathbb{E}_{x\sim\rho, (y,y')\sim \pi_s\otimes \pi_{\text{sam}}(\cdot|x)}\left[ \sum_{(+,-)}\left(\sqrt{{\mathbb{P}_{r}(y^+\succ y^-|x)}}-\sqrt{{\mathbb{P}_{r^\star}(y^+\succ y^-|x)}}\right)^2 \right]\\
    &\le 1- \frac{1}{8} \mathbb{E}_{x\sim\rho, (y,y')\sim \pi_s\otimes \pi_{\text{sam}}(\cdot|x)}\left[ \sum_{(+,-)}\left({{\mathbb{P}_{r}(y^+\succ y^-|x)}}-{{\mathbb{P}_{r^\star}(y^+\succ y^-|x)}}\right)^2 \right]\\
    &=1- \frac{1}{4} \mathbb{E}_{x\sim\rho, (y,y')\sim \pi_s\otimes \pi_{\text{sam}}(\cdot|x)}\left[ \left({{\mathbb{P}_{r}(y\succ y'|x_s)}}-{{\mathbb{P}_{r^\star}(y\succ y'|x_s)}}\right)^2 \right],
\end{align*}
where the second inequality is due to $|\sqrt{x}-\sqrt{y}|\ge |x-y|/2$ for any $x,y\in [0,1]$.
The last equality is because $|{{\mathbb{P}_{r}(y\succ y'|x_s)}}-{{\mathbb{P}_{r^\star}(y\succ y'|x_s)}}| = |{{\mathbb{P}_{r}(y'\succ y|x)}}-{{\mathbb{P}_{r^\star}(y'\succ y|x)}}|$.
Combining the above, we finally have 
\begin{align*}
    \ell(r^\star, \mathcal{D}_{t-1}) - \ell(r, \mathcal{D}_{t-1})\le -\frac{1}{2}\sum_{s=1}^{t-1} \mathbb{E}_{x\sim\rho, (y,y')\sim \pi_s\otimes \pi_{\text{sam}}(\cdot|x)}\left[ \left({{\mathbb{P}_{r}(y\succ y'|x)}}-{{\mathbb{P}_{r^\star}(y\succ y'|x)}}\right)^2 \right] +2\log \frac{|\mathcal{R}|}{\delta},
\end{align*}
which completes the proof.

\subsection{Proof of Lemma~\ref{lemma:confidence:1}}
Conditioning on the event in Lemma~\ref{lemma_MLE}, we have 
\begin{align*}
    \ell(r^\star, \mathcal{D}_{t-1}) - \ell(r, \mathcal{D}_{t-1})\le 2\log \frac{|\mathcal{R}|}{\delta}
\end{align*}
for all $r\in \mathcal{R}$ and $t\in [T]$.
This completes the proof.

\subsection{Proof of Lemma~\ref{lemma:confidence:2}}
For any $r\in \mathcal{R}$ satisfying $\ell(r,\mathcal{D}_t) - \min_{r'\in \mathcal{R}}\ell(r',\mathcal{D}_t)> 2\log \frac{|\mathcal{R}|}{\delta}$, there is
\begin{align*}
    \ell(r,\mathcal{D}_t) - \ell(r^\star, \mathcal{D}_t) > 2\log \frac{|\mathcal{R}|}{\delta}- 2\log \frac{|\mathcal{R}|}{\delta}=0
\end{align*}
since $\ell(r^\star, \mathcal{D}_{t-1}) - \min_{r'\in \mathcal{R}}\ell(r',\mathcal{D}_t)\le 2\log \frac{|\mathcal{R}|}{\delta}$ by Lemma~\ref{lemma:confidence:1}.
Therefore,
\begin{align*}
    -\ell(r, \mathcal{D}_t)+ G(r, \pi(r))I(r, \mathcal{D}_t) &= -\ell(r, \mathcal{D}_t)< -\ell(r^\star, \mathcal{D}_t)\\
    &< -\ell(r^\star, \mathcal{D}_t) + G(r^\star, \pi(r^\star))\\
    &\le \max_{r'\in \mathcal{R}}\left\{-\ell(r', \mathcal{D}_t) + G(r', \pi(r'))I(r', \mathcal{D}_t)\right\}.
\end{align*}
The last inequality is due to $I(r^\star, \mathcal{D}_t)=1$ by Lemma~\ref{lemma:confidence:1}.
This implies that such $r\not \in \mathcal{R}(\mathcal{D}_t)$ cannot be the optimal solution of \eqref{popo-obj-dpo-theory}.
In this case, it suffices to focus on $r\in \mathcal{R}(\mathcal{D}_t)$.
Consider $I(r^\star, \mathcal{D}_t)=1$ for every $r\in \mathcal{R}(\mathcal{D}_t)$, we complete the proof.

\subsection{Proof of Lemma~\ref{lemma:confidence:3}}
Let $\mathcal{F}_t$ be a filtration.
Denote by $X_t = \big(\mathbb{P}^\star(y_t\succ y_t'\mid x_t)-\mathbb{P}_r(y_t\succ y_t'\mid x_t)\big)^2$ and $P_t= \mathbb{E}_{x\sim \rho, (y,y')\sim \pi_t\otimes \pi_{\text{sam}}(\cdot|x)}\left[\left(\mathbb{P}^\star(y\succ y'\mid x)-\mathbb{P}_r(y\succ y'\mid x)\right)^2\right]$, it suffices to note that $(X_t)_{t\in \mathbb{N}^+}$ is a sequence of non-negative random variables satisfying $\mathbb{E}[X_t\mid \mathcal{F}_{t-1}] = P_t$.
We first have 
\begin{align*}
    \mathbb{E}[\exp(X_t-2P_t)\mid \mathcal{F}_{t-1}] &\le \mathbb{E}[1+(X_t-2P_t)+ (X_t-2P_t)^2\mid \mathcal{F}_{t-1}] \\
    &= \mathbb{E}[1-P_t+ X_t^2\mid \mathcal{F}_{t-1}]\le 1,
\end{align*}
where the first inequality is because $\exp(a)\le 1+a+a^2$ for $a\in [-1,1]$ and the last is due to $X_t^2\le X_t$.
Denote by $Y_t = \exp(\sum_{s=1}^t(X_s-2P_s))$, it suffices to note that $Y_1,\dots,Y_T$ is a non-negative supermartingale. 
By Ville’s inequality, we immediately have
\begin{align*}
    \mathbb{P}\left(\exists t, Y_t> \frac{1}{\delta}\right)\le \delta,
\end{align*}
which implies 
\begin{align*}
    \mathbb{P}\left(\exists t, \sum_{s=1}^t X_s> 2\sum_{s=1}^t P_s + \log\frac{1}{\delta}\right)\le \delta.
\end{align*}
Taking a union bound on $r\in \mathcal{R}$ completes the proof.

\subsection{Proof of Lemma~\ref{lemma_sigmopidlinkfunc}}
\begin{proof}
    Without loss of generality, we assume $a\ge 0$.
    We prove by case analysis.
    \begin{enumerate}
        \item ($b\in [a-1, a+1]$):
        \begin{align*}
            |\sigma(a)-\sigma(b)|\ge \dot\sigma(a+1)|a-b|\ge \frac{1}{3}\dot\sigma(a)|a-b|.
        \end{align*}
        \item ($b\not\in [a-1, a+1]$): 
        \begin{align*}
            |\sigma(a)-\sigma(b)|&= \left|\frac{1}{1+\exp(a)} - \frac{1}{1+\exp(b)}\right|\\
            &\ge \left|\frac{1}{1+\exp(a)} - \frac{1}{1+\exp(a+1)}\right| \\
            &= \frac{1}{1+\exp(a)} \frac{\exp(a+1)-\exp(a)}{1+\exp(a+1)}\\
            &\ge \frac{1}{3} \frac{1}{1+\exp(a)} \frac{\exp(a)}{1+\exp(a)} \\&\ge \frac{1}{3}\dot{\sigma}(a)\ge \frac{1}{3}\dot{\sigma}(a) \frac{|b-a|}{R_{\text{max}}}.
        \end{align*}
    \end{enumerate}
\end{proof}

\section{Generalization beyond linear preference oracle}
\label{appendix:generalization_bl}
In this section, we extend Theorem~\ref{theo:POPO} from the linear reward oracle to a more general preference oracle. 
To do this, we introduce a general complexity measure—preference-based generalized eluder coefficient (PGEC)—which aligns with the complexity measures definitions in prior works \citep{xie2024exploratory, zhang2024self}.
\begin{definition}(Preference-based GEC)
\label{def:PGEC}
Given a reward class $\mathcal{R}$, we define the preference-based Generalized Eluder Coefficient (PGEC) as the smallest $d_{\text{PGEC}}$ such that there exist $B\in O(1)$, s.t. for any $T$, $\gamma>0$, sequence of policies $\pi_t\in \Pi$ and rewards $r_t\in \mathcal{R}$ satisfying $\sum_{s=1}^{t-1} \left(\mathbb{P}^\star(y_s\succ y_s'|x_s)-\mathbb{P}_{r_t}(y_s\succ y_s'|x_s)\right)^2\le \gamma$, we have
    \begin{align*}
&\sum_{t=1}^T \mathbb{E}_{x\sim \rho, (y, y')\sim \pi_s\otimes \pi_{\text{sam}}(\cdot|x)}\left[\mathbb{P}_{r_t}\left(y\succ y'|x\right)-\mathbb{P}^\star\left(y\succ y'|x\right)  \right]\\
&\le \sqrt{d_{\text{PGEC}} \sum_{t=1}^T\sum_{s=1}^{t-1} \mathbb{E}_{x\sim \rho, (y, y')\sim \pi_s\otimes \pi_{\text{sam}}(\cdot|x)}\left[\left(\mathbb{P}_{r_t}\left(y\succ y'|x\right)-\mathbb{P}^\star\left(y\succ y'|x\right)\right)^2  \right]} +\sqrt{d_{\text{PGEC}}T}+B\gamma
\end{align*}
\end{definition}

The definition of PGEC is an variant of the Generalized Eluder Coefficient (GEC) proposed in Definition 3.4 of \cite{zhong2022gec}.
Specifically, initializing $B=\tilde{\mathcal{O}}(d\exp(R_{\text{max}}))$, it suffices to note $d_{\text{PGEC}} = d R_{\text{max}}$ for the linear reward case, as our selected $r_t$ satisfies $\sum_{s=1}^{t-1} \left(\mathbb{P}^\star(y_s\succ y_s'|x_s)-\mathbb{P}_{r_t}(y_s\succ y_s'|x_s)\right)^2\le \mathcal{O}(\log (|\mathcal{R}|/\delta))$ for every $t$.
By leveraging Definition~\ref{def:PGEC}, we can extend the proof of Theorem~\ref{theo:POPO} beyond the linear reward oracle. The only required modification is in the proof for bounding \textsc{Term 2}.
Notice that
\begin{align*}
&\textsc{Term 2}=\sum_{t=1}^T \mathbb{E}_{x\sim \rho, y\sim \pi_t(\cdot|x), y'\sim \pi_{\text{sam}}(\cdot|x)}\left[\mathbb{P}_{r_t}\left(y\succ y'|x\right)-\mathbb{P}^\star\left(y\succ y'|x\right)  \right]\\
&\le \sqrt{d_{\text{PGEC}} \sum_{t=1}^T\sum_{s=1}^{t-1} \mathbb{E}_{x\sim \rho, y\sim \pi_s(\cdot|x), y'\sim \pi_{\text{sam}}(\cdot|x)}\left[\left(\mathbb{P}_{r_t}\left(y\succ y'|x\right)-\mathbb{P}^\star\left(y\succ y'|x\right)\right)^2  \right]} +\sqrt{d_{\text{PGEC}}T} + B\gamma\\
&\le \frac{d_{\text{PGEC}}}{2\mu} + \frac{\mu}{2}\sum_{t=1}^T\sum_{s=1}^{t-1} \mathbb{E}_{x\sim \rho, y\sim \pi_s(\cdot|x), y'\sim \pi_{\text{sam}}(\cdot|x)}\left[\left(\mathbb{P}_{r_t}\left(y\succ y'|x\right)-\mathbb{P}^\star\left(y\succ y'|x\right)\right)^2  \right]+ \sqrt{d_{\text{PGEC}}T} + B\gamma,
\end{align*}
which matches the bound of \textsc{Term 2} in Theorem~\ref{theo:POPO}. 
Hence, with Definition~\ref{def:PGEC}, it suffices to say that \texttt{POPO} guarantees
\begin{align*}
    \text{Reg}_{\text{pref}}(\pi_{\text{sam}}, T)\le \tilde{{\mathcal{O}}}\left( 
\sqrt{d_{\text{PGEC}}T\log\frac{|\mathcal{R}|}{\delta}}  + \beta T  C_{\text{KL}}  \right),
\end{align*}
which also implies that the sample complexity of \texttt{SE-POPO} can be bounded by $\tilde{\mathcal{O}}\left( \frac{d_{\text{PGEC}} R_{\text{max}}^7\log\frac{|\mathcal{R}|}{\delta}}{\epsilon^2}   \right)$.

\section{Experiments Details}
\label{appendix-exp}

\subsection{Implementation Details}

\begin{algorithm}[tb]
\caption{Practical Implementation of \texttt{SE-POPO}}\label{alg:POPO-V1}
\begin{algorithmic}
   \STATE {\bfseries Input:} Reference policy $\pi_{\text{ref}}$, Prompt dataset $\mathcal{D}$, Iterations $T$
   \FOR{$t=1,\dots,T$}
   \STATE Set $\mathcal{D}_t$ as the $t$-th portion of $\mathcal{D}$ and generate $(y^1, y^2) \sim \pi_{\text{ref}}(\cdot|x)$ for each prompt $x\in \mathcal{D}_t$.
    \STATE Annotate responses $(x, y^1, y^2)\to (x, y^w, y^l)$.
    \STATE Optimize
 \begin{align*}
        \pi_{t+1} =\arg\max_{ \pi} \sum_{(x, y_w, y_l)\in \mathcal{D}_t}  \log \sigma  \bigg(\beta \log \frac{\pi(y^w|x)}{\pi_{\text{ref}}(y^w|x)} &- \beta \log \frac{\pi(y^l|x)}{\pi_{\text{ref}}(y^l|x)}\bigg) \\
        &+ \alpha  \sum_{(x, y^2)\in \mathcal{D}_t}  \sigma\bigg(- \beta \log \frac{\pi(y^2|x)}{\pi_{\text{ref}}(y^2|x)}\bigg)
    \end{align*}
    \STATE Update $\pi_{\text{ref}}\gets \pi_{t+1}$.
   \ENDFOR
\end{algorithmic}
\end{algorithm}

The experiments were conducted on 4 x Nvidia A100 80G GPUs.
The pseudocode of our algorithm's implementation is illustrated in Algorithm~\ref{alg:POPO-V1}.
In the implementation, we set $\pi_{\text{sam}}=\pi_t$ and use the chosen responses to simulate the on-policy responses. 
To accelerate training, following \cite{dong2024rlhf}, we do not restart from the initial model at each iteration but use the last-iteration model as the initial checkpoint.
Moreover, following \citet{zhang2024self}, we update $\pi_{\text{ref}}=\pi_{t+1}$ for each iteration to avoid performance regression.
For the implementations of \texttt{DPO} and \texttt{XPO}, they differ from Algorithm~\ref{alg:POPO-V1} only in the optimization objectives: \texttt{DPO} does not include the exploration bonus (i.e., $\alpha=0$), while \texttt{XPO} replaces the exploration bonus to $-\alpha \sum_{(x,y^1)\in \mathcal{D}_t} \log \frac{\pi(y^1|x)}{\pi_{\text{ref}}(y^1|x)}$.

Across all three experiments, we use \texttt{Llama-3-8B-SFT} \footnote{https://huggingface.co/RLHFlow/LLaMA3-SFT} as the base model, \texttt{RLHFlow-ultrafeedback} \footnote{https://huggingface.co/datasets/RLHFlow/ultrafeedback\_iter1, https://huggingface.co/datasets/RLHFlow/ultrafeedback\_iter2, https://huggingface.co/datasets/RLHFlow/ultrafeedback\_iter3} dataset as the training prompt sets, and \texttt{GRM-Llama3-8B-rewardmodel-ft} \footnote{https://huggingface.co/Ray2333/GRM-Llama3-8B-rewardmodel-ft} as the training preference model.
For hyperparameters, we mainly follow the settings in \cite{xie2024exploratory} and \cite{zhang2024self}.
We set $\beta=0.1$, use a global batch size of $128$, use a learning rate of $5\times 10^{-7}$ with cosine scheduling.
For exploration coefficient $\alpha$, we employ a decreasing strategy across iterations as in \cite{xie2024exploratory} and do a grid search for $\alpha$ in the first iteration over $\{0.1,0.01,0.001, 0.0001,0.00001\}$.
Based on the empirical performance on AlphcaEval benchmark, we finally select $\{1\times10^{-3}, 5\times10^{-4}, 0\}$ for \texttt{XPO} and $\{1\times10^{-1}, 5\times10^{-2}, 0\}$ for \texttt{SE-POPO} respectively.

\subsection{Academic Benchmarks}

For academic benchmarks, following \cite{xie2024exploratory}, we select tasks MMLU \cite{hendrycks2020measuring}, AGIEval \cite{zhong2023agieval}, ANLI \cite{nie2019adversarial}, GPQA \cite{rein2023gpqa}, GSM8K \cite{cobbe2021training}, WinoGrande \cite{sakaguchi2019winogrande}, TruthfulQA \cite{lin-etal-2022-truthfulqa}, ARC Challenge \cite{Clark2018ThinkYH} and HellaSwag \cite{zellers2019hellaswag} as the benchmarks.
The results are proposed in Table~\ref{tab:academic}.
It can be observed that with increasing iterations, both \texttt{SE-POPO} and other baselines may degrade on certain benchmarks, which is known as the alignment tax \cite{askell2021general, noukhovitch2024language, lin2024mitigating}.
Nevertheless, the evaluation result suggests that our method exhibits no additional degradation compared to \texttt{DPO} and \texttt{XPO}, while still effectively improving the base model across most benchmarks.

\begin{table}[h!]
    \centering
    \caption{Performance comparison across academic benchmarks}
    \label{tab:academic}
    \resizebox{1.0\linewidth}{!}{
\begin{tabular}{lccccccccc}
        \toprule
        \textbf{Model} & \textbf{MMLU} & \textbf{AGIE} & \textbf{ANLI} & \textbf{GPQA} & \textbf{GSM8K} & \textbf{WINOG} & \textbf{TRUTH} & \textbf{ARC} & \textbf{HELLA} \\
        \midrule
Llama-3-8B-SFT & 62.56 & 39.36 & 41.80 & 32.37 & 71.80 & 75.93 & 53.46 & 56.14 & 59.91\\ 
\midrule
        DPO-iter1  & 62.75 & 40.32 & 44.00 & \textbf{32.81} & 76.64 & 76.24 & 56.18 & 55.97 & 79.58 \\
        DPO-iter2  & 63.01 & 41.00 & 44.90 & 30.80 & 77.86 & 76.40 & 57.59 & 55.63 & 80.05 \\
        DPO-iter3  & 63.11 & 41.56 & \textbf{46.90} & 31.25 & 77.55 & 76.16 & \textbf{59.48} & 54.78 & 80.33 \\
        \midrule
        XPO-iter1  & 62.65 & 40.38 & 43.90 & 32.37 & 76.35 & 76.56 & 56.17 & 55.97 & 79.64 \\
        XPO-iter2  & 63.14 & 41.38 & 45.70 & 31.25 & 77.33 & 76.95 & 58.58 & 55.38 & 80.29 \\
        XPO-iter3  & 63.09 & 41.65 & 46.10 & 31.03 & \textbf{78.24} & \textbf{77.19} & 59.43 & 54.95 & 80.43 \\
        \midrule
        POPO-iter1 & 62.80 & 40.45 & 44.00 & 32.37 & 76.80 & 76.00 & 56.21 & \textbf{56.14} & 79.80 \\
        POPO-iter2 & 62.86 & 41.39 & 45.10 & 31.70 & 77.48 & 76.87 & 57.75 & 54.95 & 80.27 \\
        POPO-iter3 & \textbf{63.13} & \textbf{41.68} & 45.60 & 31.92 & 77.63 & 76.63 & 59.14 & 54.35 & \textbf{80.67} \\
        \bottomrule
    \end{tabular}  
    }
\end{table}

\newpage
\subsection{\texttt{XPO} theoretical implementation}
\begin{table*}[ht]
\centering
{%
\resizebox{1.0\linewidth}{!}{
\begin{tabular}{@{}lccccccc@{}}
\toprule
\textbf{Model} & \multicolumn{2}{c}{\textbf{IID Data}} & \multicolumn{2}{c}{\textbf{Alpaca Data}} & \multirow{2}{*}{\textbf{AE2 LC}} & \multirow{2}{*}{\textbf{MT-Bench}}& \multirow{2}{*}{\textbf{Avg. Len. (in AE2)}} \\
\cmidrule(lr){2-3} \cmidrule(lr){4-5}
 & WR & AvgR & WR & AvgR &  &  \\
\midrule
XPO-theory-iter1 & 62.6 & -4.40 & 78.3 & -5.79 & - & - & 1674\\
XPO-theory-iter2  & 68.8 & -3.37 & 87.5 & -3.79 & - & - & 1886 \\
XPO-theory-iter3 & 71.7  & -2.62 & 91.0 & -1.21 & 30.70 & 7.91 &2183 \\

\bottomrule
\end{tabular}%
}
}
\caption{Performance of \texttt{XPO} theoretical implementation}
\end{table*}

\subsection{Choices of KL-regularized coefficient $\beta$}
\begin{table*}[ht]
\centering
{%
{
\begin{tabular}{@{}lccccccc@{}}
\toprule
\textbf{Model} & \multicolumn{2}{c}{\textbf{IID Data}} & \multicolumn{2}{c}{\textbf{Alpaca Data}}   \\
\cmidrule(lr){2-3} \cmidrule(lr){4-5}
 & WR & AvgR & WR & AvgR &  &  \\
\midrule
SE-POPO-Beta-1e-1-iter1  & 62.5 & -4.32 & 80.0 & -5.68 \\
SE-POPO-Beta-1e-1-iter2  & {68.2} & -3.15 & 89.1 & -2.45 \\
SE-POPO-Beta-1e-1-iter3  & 73.3 & -2.03 & 92.4 & 0.61 \\
\midrule
SE-POPO-Beta-3e-2-iter1  & 62.3 & -4.27 & 80.6 & -5.51 \\
SE-POPO-Beta-3e-2-iter2  & 70.0 & -3.10 & 88.6 & -2.49 \\
SE-POPO-Beta-3e-2-iter3  & 72.9 & -2.01 & 93.2 & 0.83\\
\midrule
SE-POPO-Beta-1e-2-iter1  & 62.8 & -4.32 & 78.7 & -5.70 \\
SE-POPO-Beta-1e-2-iter2  & {67.5} & -3.23 & 89.4 & -2.65 \\
SE-POPO-Beta-1e-2-iter3  & 72.0 & -2.10 & 92.3 & 0.54 \\
\midrule
\bottomrule
\end{tabular}%
}
}
\caption{Performance across $\beta= \{0.1, 0.03, 0.01\}$}
\label{tab:beta}
\end{table*}


\newpage
\section*{NeurIPS Paper Checklist}

\begin{enumerate}

\item {\bf Claims}
    \item[] Question: Do the main claims made in the abstract and introduction accurately reflect the paper's contributions and scope?
    \item[] Answer: \answerYes{} 
    \item[] Justification: The abstract and introduction  accurately present the primary claims of the paper.
    \item[] Guidelines:
    \begin{itemize}
        \item The answer NA means that the abstract and introduction do not include the claims made in the paper.
        \item The abstract and/or introduction should clearly state the claims made, including the contributions made in the paper and important assumptions and limitations. A No or NA answer to this question will not be perceived well by the reviewers. 
        \item The claims made should match theoretical and experimental results, and reflect how much the results can be expected to generalize to other settings. 
        \item It is fine to include aspirational goals as motivation as long as it is clear that these goals are not attained by the paper. 
    \end{itemize}

\item {\bf Limitations}
    \item[] Question: Does the paper discuss the limitations of the work performed by the authors?
    \item[] Answer: \answerYes{} 
    \item[] Justification: The limitations are discussed in Section~\ref{sec:conclude} and the assumptions are presented in Section~\ref{sec:theory}.
    \item[] Guidelines:
    \begin{itemize}
        \item The answer NA means that the paper has no limitation while the answer No means that the paper has limitations, but those are not discussed in the paper. 
        \item The authors are encouraged to create a separate "Limitations" section in their paper.
        \item The paper should point out any strong assumptions and how robust the results are to violations of these assumptions (e.g., independence assumptions, noiseless settings, model well-specification, asymptotic approximations only holding locally). The authors should reflect on how these assumptions might be violated in practice and what the implications would be.
        \item The authors should reflect on the scope of the claims made, e.g., if the approach was only tested on a few datasets or with a few runs. In general, empirical results often depend on implicit assumptions, which should be articulated.
        \item The authors should reflect on the factors that influence the performance of the approach. For example, a facial recognition algorithm may perform poorly when image resolution is low or images are taken in low lighting. Or a speech-to-text system might not be used reliably to provide closed captions for online lectures because it fails to handle technical jargon.
        \item The authors should discuss the computational efficiency of the proposed algorithms and how they scale with dataset size.
        \item If applicable, the authors should discuss possible limitations of their approach to address problems of privacy and fairness.
        \item While the authors might fear that complete honesty about limitations might be used by reviewers as grounds for rejection, a worse outcome might be that reviewers discover limitations that aren't acknowledged in the paper. The authors should use their best judgment and recognize that individual actions in favor of transparency play an important role in developing norms that preserve the integrity of the community. Reviewers will be specifically instructed to not penalize honesty concerning limitations.
    \end{itemize}

\item {\bf Theory assumptions and proofs}
    \item[] Question: For each theoretical result, does the paper provide the full set of assumptions and a complete (and correct) proof?
    \item[] Answer: \answerYes{} 
    \item[] Justification: All of the detailed proof are provided in the appendix, including theorems, formulas, and proofs numbered and cross-referenced and assumptions stated and referenced in the statement of the theorems.
    \item[] Guidelines:
    \begin{itemize}
        \item The answer NA means that the paper does not include theoretical results. 
        \item All the theorems, formulas, and proofs in the paper should be numbered and cross-referenced.
        \item All assumptions should be clearly stated or referenced in the statement of any theorems.
        \item The proofs can either appear in the main paper or the supplemental material, but if they appear in the supplemental material, the authors are encouraged to provide a short proof sketch to provide intuition. 
        \item Inversely, any informal proof provided in the core of the paper should be complemented by formal proofs provided in appendix or supplemental material.
        \item Theorems and Lemmas that the proof relies upon should be properly referenced. 
    \end{itemize}

    \item {\bf Experimental result reproducibility}
    \item[] Question: Does the paper fully disclose all the information needed to reproduce the main experimental results of the paper to the extent that it affects the main claims and/or conclusions of the paper (regardless of whether the code and data are provided or not)?
    \item[] Answer: \answerYes{} 
    \item[] Justification: The details of experiments are in Appendix~\ref{appendix-exp}.
    \item[] Guidelines:
    \begin{itemize}
        \item The answer NA means that the paper does not include experiments.
        \item If the paper includes experiments, a No answer to this question will not be perceived well by the reviewers: Making the paper reproducible is important, regardless of whether the code and data are provided or not.
        \item If the contribution is a dataset and/or model, the authors should describe the steps taken to make their results reproducible or verifiable. 
        \item Depending on the contribution, reproducibility can be accomplished in various ways. For example, if the contribution is a novel architecture, describing the architecture fully might suffice, or if the contribution is a specific model and empirical evaluation, it may be necessary to either make it possible for others to replicate the model with the same dataset, or provide access to the model. In general. releasing code and data is often one good way to accomplish this, but reproducibility can also be provided via detailed instructions for how to replicate the results, access to a hosted model (e.g., in the case of a large language model), releasing of a model checkpoint, or other means that are appropriate to the research performed.
        \item While NeurIPS does not require releasing code, the conference does require all submissions to provide some reasonable avenue for reproducibility, which may depend on the nature of the contribution. For example
        \begin{enumerate}
            \item If the contribution is primarily a new algorithm, the paper should make it clear how to reproduce that algorithm.
            \item If the contribution is primarily a new model architecture, the paper should describe the architecture clearly and fully.
            \item If the contribution is a new model (e.g., a large language model), then there should either be a way to access this model for reproducing the results or a way to reproduce the model (e.g., with an open-source dataset or instructions for how to construct the dataset).
            \item We recognize that reproducibility may be tricky in some cases, in which case authors are welcome to describe the particular way they provide for reproducibility. In the case of closed-source models, it may be that access to the model is limited in some way (e.g., to registered users), but it should be possible for other researchers to have some path to reproducing or verifying the results.
        \end{enumerate}
    \end{itemize}

\item {\bf Open access to data and code}
    \item[] Question: Does the paper provide open access to the data and code, with sufficient instructions to faithfully reproduce the main experimental results, as described in supplemental material?
    \item[] Answer: \answerYes{} 
    \item[] Justification: We provide an anonymized version of data and code as supplemental materials.
    \item[] Guidelines:
    \begin{itemize}
        \item The answer NA means that paper does not include experiments requiring code.
        \item Please see the NeurIPS code and data submission guidelines (\url{https://nips.cc/public/guides/CodeSubmissionPolicy}) for more details.
        \item While we encourage the release of code and data, we understand that this might not be possible, so “No” is an acceptable answer. Papers cannot be rejected simply for not including code, unless this is central to the contribution (e.g., for a new open-source benchmark).
        \item The instructions should contain the exact command and environment needed to run to reproduce the results. See the NeurIPS code and data submission guidelines (\url{https://nips.cc/public/guides/CodeSubmissionPolicy}) for more details.
        \item The authors should provide instructions on data access and preparation, including how to access the raw data, preprocessed data, intermediate data, and generated data, etc.
        \item The authors should provide scripts to reproduce all experimental results for the new proposed method and baselines. If only a subset of experiments are reproducible, they should state which ones are omitted from the script and why.
        \item At submission time, to preserve anonymity, the authors should release anonymized versions (if applicable).
        \item Providing as much information as possible in supplemental material (appended to the paper) is recommended, but including URLs to data and code is permitted.
    \end{itemize}

\item {\bf Experimental setting/details}
    \item[] Question: Does the paper specify all the training and test details (e.g., data splits, hyperparameters, how they were chosen, type of optimizer, etc.) necessary to understand the results?
    \item[] Answer: \answerYes{} 
    \item[] Justification: We show the details of datasets, hyperparameters, and empirical implementation code in Appendix~\ref{appendix-exp}.
    \item[] Guidelines:
    \begin{itemize}
        \item The answer NA means that the paper does not include experiments.
        \item The experimental setting should be presented in the core of the paper to a level of detail that is necessary to appreciate the results and make sense of them.
        \item The full details can be provided either with the code, in appendix, or as supplemental material.
    \end{itemize}

\item {\bf Experiment statistical significance}
    \item[] Question: Does the paper report error bars suitably and correctly defined or other appropriate information about the statistical significance of the experiments?
    \item[] Answer: \answerNo{} 
    \item[] Justification: The experiments are conducted on models and datasets that are significantly large.
    \item[] Guidelines:
    \begin{itemize}
        \item The answer NA means that the paper does not include experiments.
        \item The authors should answer "Yes" if the results are accompanied by error bars, confidence intervals, or statistical significance tests, at least for the experiments that support the main claims of the paper.
        \item The factors of variability that the error bars are capturing should be clearly stated (for example, train/test split, initialization, random drawing of some parameter, or overall run with given experimental conditions).
        \item The method for calculating the error bars should be explained (closed form formula, call to a library function, bootstrap, etc.)
        \item The assumptions made should be given (e.g., Normally distributed errors).
        \item It should be clear whether the error bar is the standard deviation or the standard error of the mean.
        \item It is OK to report 1-sigma error bars, but one should state it. The authors should preferably report a 2-sigma error bar than state that they have a 96\% CI, if the hypothesis of Normality of errors is not verified.
        \item For asymmetric distributions, the authors should be careful not to show in tables or figures symmetric error bars that would yield results that are out of range (e.g. negative error rates).
        \item If error bars are reported in tables or plots, The authors should explain in the text how they were calculated and reference the corresponding figures or tables in the text.
    \end{itemize}

\item {\bf Experiments compute resources}
    \item[] Question: For each experiment, does the paper provide sufficient information on the computer resources (type of compute workers, memory, time of execution) needed to reproduce the experiments?
    \item[] Answer: \answerYes{} 
    \item[] Justification: Experiments compute resources are discussed in Appendix~\ref{appendix-exp}.
    \item[] Guidelines:
    \begin{itemize}
        \item The answer NA means that the paper does not include experiments.
        \item The paper should indicate the type of compute workers CPU or GPU, internal cluster, or cloud provider, including relevant memory and storage.
        \item The paper should provide the amount of compute required for each of the individual experimental runs as well as estimate the total compute. 
        \item The paper should disclose whether the full research project required more compute than the experiments reported in the paper (e.g., preliminary or failed experiments that didn't make it into the paper). 
    \end{itemize}
    
\item {\bf Code of ethics}
    \item[] Question: Does the research conducted in the paper conform, in every respect, with the NeurIPS Code of Ethics \url{https://neurips.cc/public/EthicsGuidelines}?
    \item[] Answer: \answerYes{} 
    \item[] Justification: The research conducted in the paper follows the NeurIPS Code of Ethics.
    \item[] Guidelines:
    \begin{itemize}
        \item The answer NA means that the authors have not reviewed the NeurIPS Code of Ethics.
        \item If the authors answer No, they should explain the special circumstances that require a deviation from the Code of Ethics.
        \item The authors should make sure to preserve anonymity (e.g., if there is a special consideration due to laws or regulations in their jurisdiction).
    \end{itemize}

\item {\bf Broader impacts}
    \item[] Question: Does the paper discuss both potential positive societal impacts and negative societal impacts of the work performed?
    \item[] Answer: \answerNA{} 
    \item[] Justification: There is no societal impact of the work performed.
    \item[] Guidelines:
    \begin{itemize}
        \item The answer NA means that there is no societal impact of the work performed.
        \item If the authors answer NA or No, they should explain why their work has no societal impact or why the paper does not address societal impact.
        \item Examples of negative societal impacts include potential malicious or unintended uses (e.g., disinformation, generating fake profiles, surveillance), fairness considerations (e.g., deployment of technologies that could make decisions that unfairly impact specific groups), privacy considerations, and security considerations.
        \item The conference expects that many papers will be foundational research and not tied to particular applications, let alone deployments. However, if there is a direct path to any negative applications, the authors should point it out. For example, it is legitimate to point out that an improvement in the quality of generative models could be used to generate deepfakes for disinformation. On the other hand, it is not needed to point out that a generic algorithm for optimizing neural networks could enable people to train models that generate Deepfakes faster.
        \item The authors should consider possible harms that could arise when the technology is being used as intended and functioning correctly, harms that could arise when the technology is being used as intended but gives incorrect results, and harms following from (intentional or unintentional) misuse of the technology.
        \item If there are negative societal impacts, the authors could also discuss possible mitigation strategies (e.g., gated release of models, providing defenses in addition to attacks, mechanisms for monitoring misuse, mechanisms to monitor how a system learns from feedback over time, improving the efficiency and accessibility of ML).
    \end{itemize}
    
\item {\bf Safeguards}
    \item[] Question: Does the paper describe safeguards that have been put in place for responsible release of data or models that have a high risk for misuse (e.g., pretrained language models, image generators, or scraped datasets)?
    \item[] Answer: \answerNA{} 
    \item[] Justification: The paper poses no such risks.
    \item[] Guidelines:
    \begin{itemize}
        \item The answer NA means that the paper poses no such risks.
        \item Released models that have a high risk for misuse or dual-use should be released with necessary safeguards to allow for controlled use of the model, for example by requiring that users adhere to usage guidelines or restrictions to access the model or implementing safety filters. 
        \item Datasets that have been scraped from the Internet could pose safety risks. The authors should describe how they avoided releasing unsafe images.
        \item We recognize that providing effective safeguards is challenging, and many papers do not require this, but we encourage authors to take this into account and make a best faith effort.
    \end{itemize}

\item {\bf Licenses for existing assets}
    \item[] Question: Are the creators or original owners of assets (e.g., code, data, models), used in the paper, properly credited and are the license and terms of use explicitly mentioned and properly respected?
    \item[] Answer: \answerYes{} 
    \item[] Justification: We cite the models and dataset used in the paper in Appendix~\ref{appendix-exp}.
    \item[] Guidelines:
    \begin{itemize}
        \item The answer NA means that the paper does not use existing assets.
        \item The authors should cite the original paper that produced the code package or dataset.
        \item The authors should state which version of the asset is used and, if possible, include a URL.
        \item The name of the license (e.g., CC-BY 4.0) should be included for each asset.
        \item For scraped data from a particular source (e.g., website), the copyright and terms of service of that source should be provided.
        \item If assets are released, the license, copyright information, and terms of use in the package should be provided. For popular datasets, \url{paperswithcode.com/datasets} has curated licenses for some datasets. Their licensing guide can help determine the license of a dataset.
        \item For existing datasets that are re-packaged, both the original license and the license of the derived asset (if it has changed) should be provided.
        \item If this information is not available online, the authors are encouraged to reach out to the asset's creators.
    \end{itemize}

\item {\bf New assets}
    \item[] Question: Are new assets introduced in the paper well documented and is the documentation provided alongside the assets?
    \item[] Answer: \answerYes{} 
    \item[] Justification: The code used in the paper is well documented with instructions.
    \item[] Guidelines:
    \begin{itemize}
        \item The answer NA means that the paper does not release new assets.
        \item Researchers should communicate the details of the dataset/code/model as part of their submissions via structured templates. This includes details about training, license, limitations, etc. 
        \item The paper should discuss whether and how consent was obtained from people whose asset is used.
        \item At submission time, remember to anonymize your assets (if applicable). You can either create an anonymized URL or include an anonymized zip file.
    \end{itemize}

\item {\bf Crowdsourcing and research with human subjects}
    \item[] Question: For crowdsourcing experiments and research with human subjects, does the paper include the full text of instructions given to participants and screenshots, if applicable, as well as details about compensation (if any)? 
    \item[] Answer: \answerNA{} 
    \item[] Justification: The paper does not involve crowdsourcing nor research with human subjects.
    \item[] Guidelines:
    \begin{itemize}
        \item The answer NA means that the paper does not involve crowdsourcing nor research with human subjects.
        \item Including this information in the supplemental material is fine, but if the main contribution of the paper involves human subjects, then as much detail as possible should be included in the main paper. 
        \item According to the NeurIPS Code of Ethics, workers involved in data collection, curation, or other labor should be paid at least the minimum wage in the country of the data collector. 
    \end{itemize}

\item {\bf Institutional review board (IRB) approvals or equivalent for research with human subjects}
    \item[] Question: Does the paper describe potential risks incurred by study participants, whether such risks were disclosed to the subjects, and whether Institutional Review Board (IRB) approvals (or an equivalent approval/review based on the requirements of your country or institution) were obtained?
    \item[] Answer: \answerNA{} 
    \item[] Justification: The paper does not involve crowdsourcing nor research with human subjects.
    \item[] Guidelines:
    \begin{itemize}
        \item The answer NA means that the paper does not involve crowdsourcing nor research with human subjects.
        \item Depending on the country in which research is conducted, IRB approval (or equivalent) may be required for any human subjects research. If you obtained IRB approval, you should clearly state this in the paper. 
        \item We recognize that the procedures for this may vary significantly between institutions and locations, and we expect authors to adhere to the NeurIPS Code of Ethics and the guidelines for their institution. 
        \item For initial submissions, do not include any information that would break anonymity (if applicable), such as the institution conducting the review.
    \end{itemize}

\item {\bf Declaration of LLM usage}
    \item[] Question: Does the paper describe the usage of LLMs if it is an important, original, or non-standard component of the core methods in this research? Note that if the LLM is used only for writing, editing, or formatting purposes and does not impact the core methodology, scientific rigorousness, or originality of the research, declaration is not required.
    \item[] Answer: \answerNA{} 
    \item[] Justification: The core method development in this research does not involve LLMs as any important, original, or non-standard components.
    \item[] Guidelines:
    \begin{itemize}
        \item The answer NA means that the core method development in this research does not involve LLMs as any important, original, or non-standard components.
        \item Please refer to our LLM policy (\url{https://neurips.cc/Conferences/2025/LLM}) for what should or should not be described.
    \end{itemize}

\end{enumerate}

\end{document}